# TCMClinicalReason-Bench: Can Language Models Reason from Pathogenesis to Prescription over Real-World Clinical Cases?

**Jirui Dai[3,1], Chenkai Zhang[6,*], Yan Jia[4,*], Yukai Wang[6], Ruiyang He[5], Changyong Luo[2,†], Zhi Liu[1,2]**

**Affiliation**

[1] the School of Pharmacy, Nanjing University of Chinese Medicine, Nanjing, China

[2] the Infectious disease department, Dongfang Hospital, Beijing University of Chinese Medicine, Beijing, China

[3] the Department of Computer Science, Johns Hopkins University, Baltimore, USA

[4] Graduate School, the Beijing University of Chinese Medicine, Beijing, China

[5] Dongzhimen Hospital, Beijing University of Chinese Medicine, Beijing, China

[6] the School of Information Engineering, Huzhou University, Huzhou, China

## Abstract

Large language models (LLMs) can generate clinical narratives that are insufficiently grounded in patient-specific evidence. In traditional Chinese medicine (TCM), errors can propagate from etiology and pathogenesis through syndrome diagnosis and treatment principles to prescription generation. We developed TCMClinicalReason-Bench using 2,000 multicenter electronic health record cases to distinguish case-grounded responses from fluent but unsupported diagnostic and therapeutic conclusions. Five general-purpose and two TCM-specific LLMs were evaluated in zero-shot settings. An evidence-constrained rubric assessed seven diagnostic and therapeutic components and three cross-block relations, allowing case-supported alternatives. Qwen3.7-Plus with TCM retrieval served as the automated judge, alongside parallel blinded ratings by five senior TCM clinicians on a 600-case subset. Structural completeness was nearly saturated (99.3 – 100.0%), but normalized content scores ranged from 40.7% to 54.1%. The five general-purpose models averaged 50.0%, versus 41.3% for the two smaller TCM-specific models. Cross-block logic consistency ranged from 60.8% to 66.8% and correlated moderately with content across cases (Pearson’s r = 0.515 – 0.656). Deficits were greatest in prescription generation, prescription analysis, and symptom-guided modification. In judge stress testing on 100 independent cases, perturbation detection rates across the three relations were 57%, 56%, and 31%, with contradictions detected more reliably than omissions. Separating component quality from cross-block consistency localizes failures missed by endpoint and completeness metrics and identifies where clinician oversight remains necessary.

***Keywords*** Medical AI benchmark, Clinical Reasoning, Large Language Models, Clinical decision-making, Evidence-grounded Evaluation.

## Introduction

Traditional Chinese medicine（TCM） remains an important branch of traditional medicine, supported by a systematic theoretical framework and extensive clinical practice[1,2]. Its core process of syndrome differentiation and treatment, however, is sequential, individualized, and evidence-dependent[3,4]. Clinicians integrate symptoms, physical signs, medical history, and tongue and pulse findings to infer etiology and pathogenesis, identify the syndrome; formulate treatment principles, select and explain a prescription, and modify treatment and prescription as the patient's condition evolves. **TCM clinical competence therefore cannot be reduced to producing a diagnosis or prescription in isolation.** Each decision constrains the next, although more than one pathway can be clinically defensible when supported by the same case evidence. Mastering this process requires prolonged professional training and clinical experience[5]. Large language models (LLMs) may help synthesize heterogeneous records and domain knowledge[6,7], integrate complex clinical information[8,9], and support more accessible, evidence-informed TCM decision-making[6,10]. Their clinical value, however, depends on whether patient-specific evidence is preserved across the complete diagnostic and therapeutic procedure. This raises the central concern: **Can LLMs reason from pathogenesis to prescription in real-world cases while grounding in patient evidence and maintaining coherence across stages, and distinguish clinically grounded reasoning from narratives that are internally coherent but weakly supported by the case?**

---

[1] The first three authors contribute the equally contributions, and should regarded as co-first authors.

[2] Corresponding authors: Zhi Liu (zhiliu@njucm.edu.cn), Changyong Luo (Bdf01344@bucm.edu.cn)

Fluent clinical language makes this requirement difficult to evaluate. LLMs may produce coherent and authoritative-sounding rationales even when their conclusions are unsupported[11]. hallucinate information, contradict established medical knowledge, or generate internally inconsistent recommendations[12,13]. These risks are particularly salient in TCM, where an error in etiology, pathogenesis, or syndrome identification can propagate through the treatment principle to an executable prescription[14] and may have safety implications[15]. Evaluation based only on a final diagnosis, final prescription, lexical overlap, apparent coherence, or output completeness therefore collapses two distinct questions: whether each clinical decision is supported by the case and whether successive decisions are clinically connected. It may also penalize a defensible alternative pathway while rewarding superficial agreement with a single reference answer. Authentic cases and explicit cross-component assessment are consequently needed to examine both component-level adequacy and cross-stage reasoning consistency[16].

Existing medical LLM benchmarks broadly address three related but incomplete evaluation settings. Knowledge- and examination-oriented resources, including MedXpertQA, CMB, and MedReason, primarily assess medical knowledge or structured reasoning against predefined targets[17-19]. Open-ended frameworks such as HealthBench use physician-authored rubrics to evaluate realistic health conversations [20]. TCM-specific benchmarks, including TCMEval-SDT[21], TCM-3CEval[22], MTCMB[23], TCM-BEST4SDT[24], and DCB[25], have extended evaluation to syndrome-related tasks, clinical decision-making, prescriptions, safety, dynamic workflows, and human review; case-based studies have also compared model-generated diagnostic and treatment recommendations with clinical experts[26]. Collectively, these studies have substantially advanced medical and TCM evaluation, but they generally emphasize final answers, aggregate performance, or selected stages of clinical decision-making. Joint assessment of detailed component adequacy and prespecified dependencies within the same open-ended case response remains limited. In particular, few benchmarks are designed to determine where patient evidence is lost along a multistage diagnostic and therapeutic pathway, distinguish a case-supported alternative pathway from an unsupported deviation, and characterize a scalable automated judge using both controlled error injection and blinded clinician ratings. The central gap is therefore not simply the absence of a larger TCM dataset or another measure of response completeness. It is the absence of a clinically interpretable framework that separately evaluates whether individual decisions are grounded in patient evidence and whether those decisions form a coherent pathway from pathogenesis to prescription.

To address this gap, we developed a clinically grounded benchmark of 2000 cases spanning 443 recorded disease labels, curated from multicenter electronic health records, and evaluated five general-purpose and two TCM-specific LLMs. The framework measures two complementary layers: Block-level evaluation assesses seven open-ended components: syndrome-differentiation and treatment reasoning, etiology and pathogenesis, syndrome diagnosis, treatment-principles selection, prescription generation, prescription analysis, and prescription modification in response to changes in symptoms. Cross-block evaluation separately examines three core dependency pathways: etiology and pathogenesis to syndrome diagnosis, syndrome diagnosis to treatment-principles selection, and treatment-principles selection to prescription generation and prescription analysis. The evidence-constrained rubric permits case-supported alternative pathways but requires explicit supporting evidence, counterevidence, a score, and a rationale.

The automated evaluation architecture was developed in two stages with distinct purposes. First, a **judge model selection evaluation was first conducted** and selected Qwen3.7-Plus as the judge model through a retrieval-free preliminary comparison of candidate judge configurations. Then, an independent test set with 100 clinical cases were constructed and performed over the three cross-block reasoning pathways. This perturbation experiment further tested the integrated architecture's sensitivity to injected logical errors rather than isolating the incremental effect of TCM-RAG. Second, **a standardized TCM retrieval-augmented generation tool was then incorporated for formal benchmark scoring**. After that, an independent, blinded clinician ratings provide evidence for the clinical validity of the automated evaluation and help identify systematic disagreements or failure modes, which delineates the range of clinically acceptable reasoning and retain responsibility for adjudicating ambiguous, unconventional, or safety-relevant cases. Under this human–AI collaborative design, clinical experts define the evaluation constructs, scoring criteria, and boundaries of acceptable diagnostic and therapeutic reasoning, whereas the LLM-based judge operationalizes these criteria at scale, enabling consistent, fine-grained, and auditable assessment across the full benchmark. The results showed a clear gap between structural completeness and clinical substance in formal evaluation. Although nearly all required sections were present (99.3%–100.0%), normalized block-level content scores ranged from only 40.7% to 54.1%. Within the evaluated set, the five general-purpose models had a higher descriptive mean than the two TCM-specific models, suggesting that domain specialization alone did not ensure stronger case-level performance. Deficits were greatest in prescription generation, prescription analysis, and symptom-guided modification, while the treatment-principle–prescription–analysis relation was the weakest cross-block dependency. Cross-block consistency varied less (60.8%–66.8%) and correlated only moderately with content scores ($r = 0.515$–$0.656$), whereas content declined more sharply with difficulty. These patterns were not materially altered by source stratification or contamination identification. Perturbation testing further showed that explicit contradictions were detected more reliably than omissions or subtle prescription-analysis errors.

These findings distinguish procedural fluency from case-grounded clinical competence. By locating failures both within individual components and at the transitions between them, the benchmark provides an auditable basis for model development and clinician oversight. Improvement should focus on retaining patient evidence throughout the clinical pathway, aligning treatment principles with prescriptions, making prescription rationales traceable, and defining explicit triggers for follow-up modification. The benchmark may extend to other TCM specialties, traditional medicine systems, and multistage clinical tasks, but its transferability requires independent external validation. It also needs to notice that this benchmark is intended to characterize model capabilities and failure modes in clinical healthcare rather than to establish a threshold for autonomous clinical use.

# Experiments

## The Design of Judge Model Selection

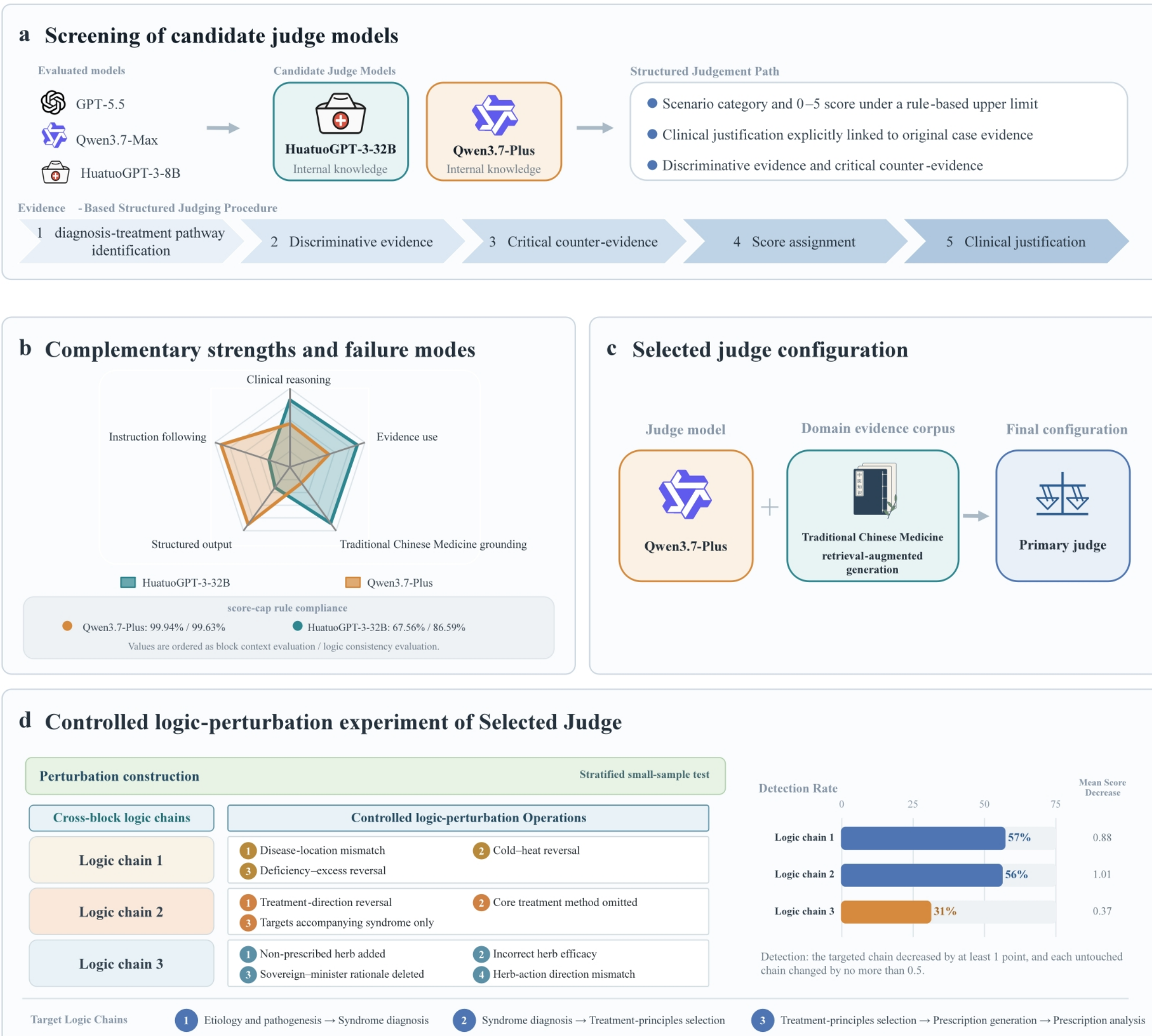

**Fig.1. Selection and validation of the primary LLM judge. a**, HuatuoGPT-3-32B and Qwen3.7-Plus evaluated model responses using a structured protocol covering pathway classification, case evidence, counterevidence, capped 0–5 scoring, and clinical justification. **b**, The candidates were compared for clinical reasoning, instruction following, evidence use, structured output, TCM grounding, and score-cap compliance. **c**, Qwen3.7-Plus with TCM retrieval-augmented generation was selected. **d**, Sensitivity to reasoning errors was tested by an independent test set for three cross-block reasoning pathways. Detection rates were 57%, 56%, and 31%, with mean score decreases of 0.88, 1.01, and 0.37 points. The controlled perturbation experiment was a separate post-selection stress test of judge sensitivity and was not part of the clinician–judge comparison.

## Structured Evaluation Protocol and Candidate Judges

Before the full benchmark evaluation, we conducted a preliminary study to determine whether candidate LLM judges could reliably assess TCM clinical reasoning ability of LLMs. The analysis covered two task sets from complementary perspectives. The block-level assessment evaluated correctness, completeness, and clinical appropriateness across seven tasks: *syndrome-differentiation-and-treatment reasoning path*, *etiology and pathogenesis analysis*, *syndrome diagnosis*, *treatment-principle selection*, *prescription generation*, *prescription analysis*, and *prescription modification in response to changes in symptoms*. The cross-block assessment examined whether each response formed a coherent, case-supported reasoning chain from etiology and pathogenesis to syndrome diagnosis, from syndrome diagnosis to treatment-principles selection, and from treatment-principles selection through prescription generation to prescription analysis. Fig.1 shows the process of Judge model selection.

TCM diagnosis permits more than one clinically defensible treatment pathway, consistent with the principle that the same disease may require different treatments. Reference-answer matching alone is therefore insufficient. A response may follow the same core diagnosis-treatment reasoning as the reference or present an alternative pathway supported by case evidence. To distinguish a defensible alternative from a fluent but unsupported account, we implemented an evidence-based structured judging protocol. Each judge assigned (i) a diagnosis-treatment pathway identification, (ii) case-specific discriminating evidence, (iii) the strongest counterevidence, (iv) a score, and (v) a rationale. Scores were assigned on a common 0–5 scale with scenario-dependent caps. Supporting evidence had to be drawn from the source case, whereas the counterevidence assessment tested for unresolved conflicts among syndrome diagnosis, treatment-principles selection, prescription generation, and prescription analysis.

The candidate judges were HuatuoGPT-3-32B and Qwen3.7-Plus[3]. HuatuoGPT-3-32B served as the domain-specific candidate and was a comparatively large, deployable TCM-focused model. It was included to examine the performance of a domain-specialized judge when evaluating symptoms, tongue and pulse findings, pathogenesis, prescription, materia medica, and case evidence. Qwen3.7-Plus served as the general-purpose candidate because of its capacity for long-context processing, complex instruction following, and structured output. We refer to them as the Huatuo judge and the Qwen judge, respectively, to distinguish them from the evaluated models: GPT-5.5, Qwen3.7-Max, and HuatuoGPT-3-8B.

**Preliminary Evaluation Dataset Composition and Analysis Plan**

The preliminary study for Judge LLM selection used an independent subset with 462 samples that remained in the pool of 6,123 cases that was not selected in the benchmark. Each case included responses from GPT-5.5, Qwen3.7-Max, and HuatuoGPT-3-8B, as shown in Table 1.

**Table 1. Composition of the preliminary-study dataset and evaluable samples.**

| Item | Composition | Analytical purpose |
|---|---|---|
| Cases | 462 cases | The distribution is strongly long-tailed; disease-stratified analyses require a minimum sample-size threshold |
| Case difficulty | 154 hard; 154 medium; 154 easy | The strata support assessment of sensitivity to case complexity |
| Evaluated models | GPT-5.5; Qwen3.7-Max; HuatuoGPT-3-8B | Selected to span two general-purpose model families and one TCM-specific model family |
| Candidate judges | Qwen3.7-Plus; HuatuoGPT-3-32B | Selected to compare a general-purpose judge with a TCM-specialized judge |

The preliminary study was not designed to establish either candidate judge as a clinical gold standard. Its purpose was to select a judge configuration that was sufficiently stable, auditable, and compatible with domain-specific evidence retrieval for the formal benchmark. We evaluated each candidate using four observable criteria. **Procedural reliability** assessed complete case coverage, valid and complete outputs, and consistent application of the score-cap rule[4]. **Construct validity** was examined by determining whether the scores distinguished patterns in model ordering, case difficulty, and the sensitivity to artificially introduced logical errors. **Discriminative capacity** measured whether the judge separated model outputs within the same case without excessive clustering at the upper end of the scale. **Clinical coherence** assessed whether the diagnosis-treatment pathway category, supporting evidence, counterevidence, score, and rationale were mutually consistent and clinically defensible. The selected judge configuration was subsequently evaluated using controlled logical perturbation study for error-sensitivity validation.

[3] Qwen3.7-Plus was selected on the basis of access stability, inference cost, structured-output availability, and overall performance among general-purpose models. Its selection does not imply that other state-of-the-art general-purpose models lack comparable judging capability.

[4] For each evaluated unit, the judge assigned a raw score from 0 to 5 and classified the response as aligned with the core reasoning of the reference answer, a valid alternative pathway, or an unsupported alternative pathway. These categories imposed score caps of 5, 4, and 3, respectively. The score used in the analysis was the lower of the raw score and the applicable cap. A response initially classified as a valid alternative was downgraded to an unsupported alternative, with a maximum score of 3, if the judge supplied no case-specific discriminating evidence or identified unresolved critical counterevidence.

## Preliminary Evaluation on Judge Model Selection

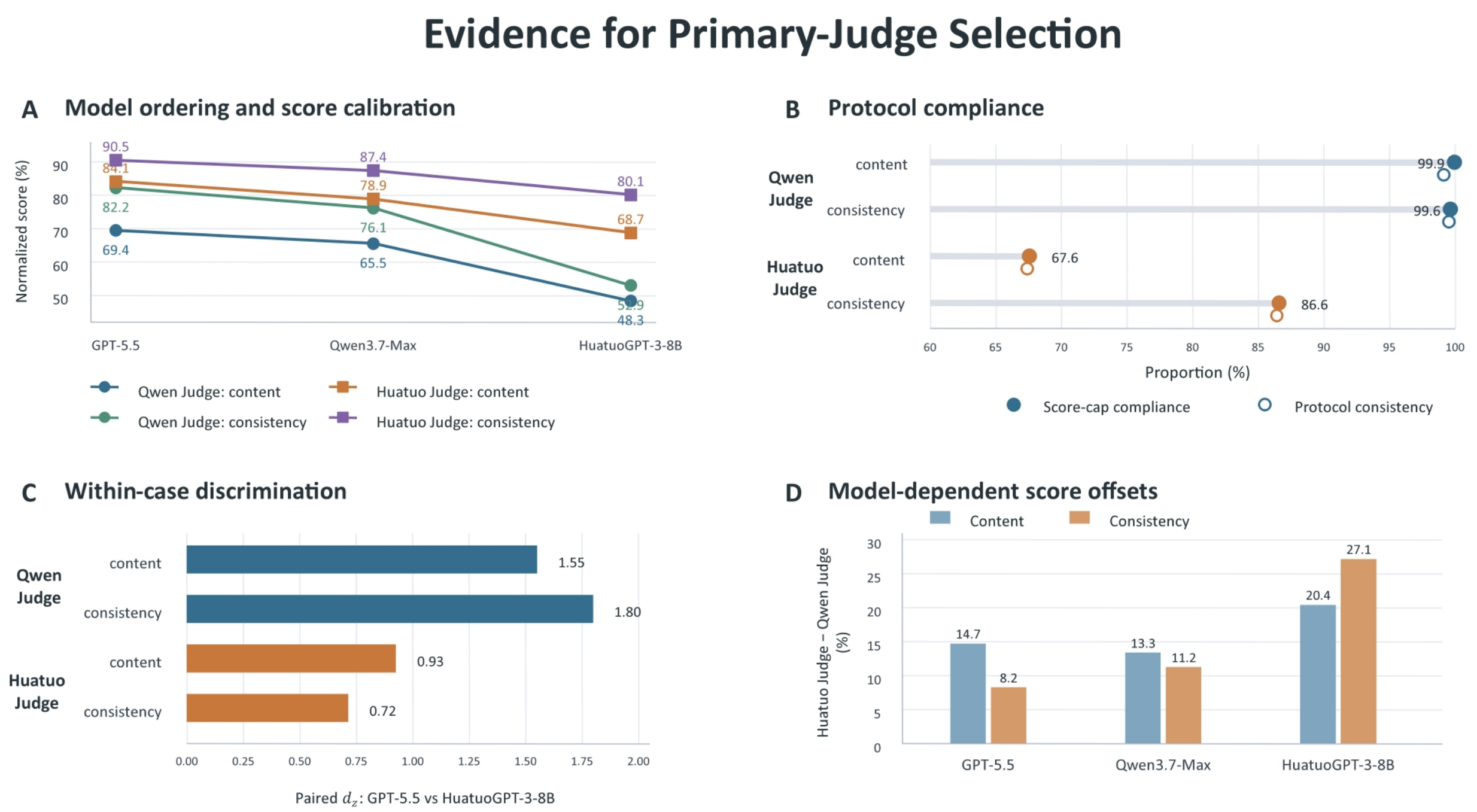

**Fig.2.** Evidence used to select the Judge Model. (A) Aggregate model ordering and normalized scores; (B) score-cap compliance and case-level assessment completion; (C) standardized within-case effect size ($d_z$) for GPT-5.5 versus HuatuoGPT-3-8B responses; and (D) mean score difference (Huatuo judge minus Qwen judge) across evaluated models, used to assess potential same-family preference.

## Overall Scoring Behavior and Procedural Reliability

Fig.2 summarizes four quantitative comparisons of the candidate judges. Panel A reports aggregate model ordering and differences in scoring scale; panel B summarizes rule compliance and case-level assessment completion; panel C quantifies within-case discrimination; and panel D examines whether the score difference between judges was disproportionately large for Huatuo-generated responses.

Both judges produced the same aggregate ordering: GPT-5.5 > Qwen3.7-Max > HuatuoGPT-3-8B. Their scoring scales, however, differed substantially. In the content assessment, the Qwen judge assigned mean normalized scores of 69.4%, 65.5%, and 48.3% to the three models, respectively; the corresponding cross-block logic consistency scores were 82.2%, 76.1%, and 52.9%. The Huatuo judge assigned content scores of 84.1%, 78.9%, and 68.7% and consistency scores of 90.5%, 87.4%, and 80.1%. The Huatuo judge assigned higher scores across all models, and the largest between-judge difference was observed for HuatuoGPT-3-8B. Thus, the between-judge difference could not be explained by a simple uniform upward shift in scores.

Procedural reliability provided a clearer distinction between the two candidates. The Qwen judge returned complete content and consistency assessments for all test cases; the Huatuo judge omitted three cases in the content assessment and one in the consistency assessment. Score-cap compliance was 99.94% and 99.63% for the Qwen judge, compared with 67.56% and 86.59% for the Huatuo judge. Because the cap rule directly links pathway classification to the permissible score, these violations demonstrate inconsistent rule application. The higher mean scores may therefore reflect both differences in score calibration and differences in rule compliance.

### Within-Case Discrimination, Difficulty Sensitivity, and Potential Same-Family Preference

We assessed within-case discrimination using paired GPT-5.5 and HuatuoGPT-3-8B responses for each case. As shown in panel C, the Qwen judge yielded standardized paired effect sizes ($d_z$) of 1.55 and 1.80 for the content and consistency assessments, respectively; the corresponding values for the Huatuo judge were 0.93 and 0.72. Both judges separated the model outputs in the expected direction, but the Qwen judge produced greater standardized separation between them.

We then examined whether the between-judge offset varied by evaluated model. Under a uniform calibration shift, the Huatuo-minus-Qwen difference would be comparable for GPT, Qwen, and Huatuo outputs. Instead, the mean differences in the content assessment were 14.7, 13.4, and 20.4 percentage points for GPT-5.5, Qwen3.7-Max, and HuatuoGPT-3-8B, respectively (Fig.2-D). The additional increase for Huatuo outputs relative to the other two evaluated model families was 6.35 percentage points (95% bootstrap CI, 5.74–6.98). In the consistency assessment, the corresponding excess between-judge difference was 17.39 percentage points (95% CI, 16.66–18.13). The Huatuo judge also classified fewer Huatuo responses as unsupported alternative pathways and assigned a larger proportion of scores between 4 and 5. This model-dependent divergence is compatible with, but does not establish, a same-family scoring preference because no independent clinical reference was used to adjudicate the judges.

**Table 2. Normalized scores (0–1) by case difficulty and easy–hard differences**

| Judge | Assessment | Evaluated model | Hard | Medium | Easy | Easy − hard difference (%) |
|---|---|---|---|---|---|---|
| Qwen3.7-Plus | Block-level content | GPT-5.5 | 0.647 | 0.693 | 0.743 | 9.6 |
| | | Qwen3.7-Max | 0.613 | 0.658 | 0.696 | 8.3 |
| | | HuatuoGPT-3-8B | 0.429 | 0.476 | 0.546 | 11.7 |
| | Cross-block logic consistency | GPT-5.5 | 0.789 | 0.823 | 0.854 | 6.5 |
| | | Qwen3.7-Max | 0.732 | 0.762 | 0.791 | 5.9 |
| | | HuatuoGPT-3-8B | 0.498 | 0.521 | 0.570 | 7.2 |
| HuatuoGPT-3-32B | Block-level content | GPT-5.5 | 0.789 | 0.846 | 0.889 | 10.0 |
| | | Qwen3.7-Max | 0.748 | 0.789 | 0.829 | 8.1 |
| | | HuatuoGPT-3-8B | 0.626 | 0.678 | 0.758 | 13.2 |
| | Cross-block logic consistency | GPT-5.5 | 0.871 | 0.903 | 0.939 | 6.8 |
| | | Qwen3.7-Max | 0.845 | 0.872 | 0.904 | 5.9 |
| | | HuatuoGPT-3-8B | 0.766 | 0.792 | 0.845 | 7.9 |

For every evaluated model, as shown in Table 2, scores increased monotonically from hard to medium to easy cases. The between-judge offset nevertheless persisted across all three strata and remained largest for HuatuoGPT-3-8B. In the consistency assessment, for example, the Huatuo-minus-Qwen difference for HuatuoGPT-3-8B was 26.8 percentage points in hard cases and 27.5 points in easy cases.

### Inter-Judge Agreement and Its Interpretation

Agreement analyses were used to determine whether the two judges preserved similar case-level orderings and to characterize systematic score offsets; agreement itself was not interpreted as evidence of clinical accuracy. In this analysis, Spearman's ρ was used to measure case-level rank association between the two judges. Spearman's ρ values for case-level mean scores ranged from 0.668 to 0.689 for the content assessment and from 0.351 to 0.465 for the consistency assessment. Thus, rank association was higher for content scores than for consistency scores. Lin's concordance correlation coefficients (CCCs) were lower, ranging from 0.303 to 0.408 for content and from 0.125 to 0.359 for consistency, indicating limited absolute-score concordance, particularly for cross-block clinical reasoning consistency.

Disagreement was greatest for HuatuoGPT-3-8B responses. For this model, the CCC was 0.303 in the content assessment and 0.125 in the consistency assessment, and the proportion of cases in which the judges differed by no more than one raw score point was 47.6% and 30.3%, respectively. For the treatment-principles selection through prescription generation to prescription analysis relation, quadratic weighted $\kappa$ was 0.075, indicating minimal ordinal agreement beyond chance. These estimates are consistent with the model-dependent score offset shown in Fig.2-D. Detailed results are reported in **Appendix G**.

**Validation Using Controlled Logical Perturbations**

Before the selected judge configuration was used in the formal block-level and cross-block logic consistency evaluations, we conducted a controlled perturbation study to assess its sensitivity to cross-block clinical reasoning errors. We collect 100 clinical cases which are independent from TCMClinicalReason-Bench with the difficulty-labeled test set (34 hard, 33 medium, and 33 easy). Responses generated by GPT-5.5, Qwen3.7-Max, and HuatuoGPT-3-8B form 300 original responses. Each response was perturbed once for each of the three cross-block reasoning pathways, producing 900 matched original–perturbed pairs (300 per reasoning pathways).

***Perturbation construction***. Perturbations were generated with DeepSeek-V4-Pro under a controlled rewriting protocol. Each variant introduced one predefined defect in a single target relation. We preserved case information, non-target content, writing style, response length, and domain terminology wherever possible. Minimal downstream edits were permitted only when needed to avoid additional inconsistencies outside the target relation. This design was intended to attribute score changes to the injected defect rather than to differences in surface form.

For the cross-block logic consistency of etiology and pathogenesis to syndrome diagnosis, only the syndrome diagnosis was altered so that it was no longer supported by the stated pathogenesis[5]. Perturbations introduced a disease-location mismatch, a cold–heat reversal, or a deficiency–excess reversal; the pathogenesis and downstream blocks were otherwise retained. For the relation between syndrome diagnosis and treatment principle, the treatment principle was altered by reversing the treatment direction, omitting the core treatment, or addressing only a secondary syndrome. Prescription generation and prescription analysis were minimally adjusted to remain internally consistent with the modified treatment principle, thereby restricting the defect to the relation between syndrome diagnosis and treatment principle. For the relation linking treatment principle, prescription generation, and prescription analysis, only the prescription analysis was rewritten. Perturbations referred to a medicinal absent from the prescription, assigned an incorrect action to a prescribed medicinal, omitted the monarch–minister rationale while retaining the remaining explanation, or inferred from the prescribed medicinal a therapeutic direction that contradicted the stated treatment principle.

---

[5] This relation was selected because it captures the key transition from explanatory reasoning to diagnostic determination in TCM. Etiology and pathogenesis analysis synthesizes the patient's manifestations into an account of the current pathological state, whereas syndrome diagnosis translates that account into a clinically actionable characterization that provides the basis for subsequent treatment-principle selection and prescription planning. Errors at this interface may therefore redirect the entire downstream therapeutic pathway, even when the subsequent components appear internally coherent, making it a clinically consequential target for controlled perturbation.

**Table 3.** Sensitivity to controlled logical perturbations in 900 matched pairs from 100 cases. Results are pooled across the three evaluated models. Original and perturbed scores are primary-judge ratings on a 0–5 scale. Δ is defined as perturbed minus original score and is reported as mean ± SD. A perturbation was detected when the targeted relation decreased by at least 1 point and the absolute change in each untargeted relation did not exceed 0.5. Non-targeted content was held constant except for minimal downstream edits required to avoid additional inconsistencies.

| Targeted relation | Perturbation type | Original score, mean | Perturbed score, mean | Score change, Δ (mean ± SD) | Detection rate |
|---|---|---|---|---|---|
| Etiology and pathogenesis → syndrome diagnosis | Disease-location mismatch | 3.15 | 2.32 | −0.82 ± 0.70 | 60% |
| | Cold–heat reversal | 3.12 | 1.91 | −1.21 ± 0.91 | 71% |
| | Deficiency–excess reversal | 3.18 | 2.58 | −0.61 ± 0.73 | 40% |
| | Overall | 3.15 | 2.27 | −0.88 ± 0.82 | 57% |
| Syndrome diagnosis → treatment principle | Treatment-direction reversal | 3.24 | 1.76 | −1.47 ± 0.92 | 69% |
| | Core treatment principle omitted | 3.20 | 2.45 | −0.75 ± 0.79 | 44% |
| | Treatment directed only at a secondary syndrome | 3.27 | 2.46 | −0.81 ± 0.72 | 56% |
| | Overall | 3.24 | 2.22 | −1.01 ± 0.88 | 56% |
| Treatment principle → prescription → prescription analysis | Non-prescribed medicinal substance discussed | 3.05 | 2.80 | −0.25 ± 0.52 | 24% |
| | Incorrect attribution of a medicinal substance's action | 3.04 | 2.11 | −0.93 ± 0.70 | 71% |
| | Monarch–minister rationale omitted | 3.04 | 2.99 | −0.05 ± 0.32 | 8% |
| | Mismatch between attributed medicinal actions and treatment direction | 3.03 | 2.77 | −0.25 ± 0.52 | 23% |
| | Overall | 3.04 | 2.67 | −0.37 ± 0.63 | 31% |

Results are demonstrated in Table 3. Sensitivity varied substantially by perturbation type. Explicit contradictions were detected more reliably: treatment-direction reversal was detected in 69% of pairs (Δ = −1.47), cold–heat reversal in 71% (Δ = −1.21), and incorrect attribution of a medicinal substance's action in 71% (Δ = −0.93). Mean signed changes in the untargeted pathways were within ±0.12. This aggregate result suggests limited spillover but does not exclude offsetting case-level changes. Omission-based or structural errors were detected less often. Omission of the monarch–minister rationale was detected in 8% of pairs (Δ = −0.05), reference to a non-prescribed medicinal substance in 24%, and a mismatch between attributed medicinal actions and the inferred treatment direction in 23%. A similar pattern was observed in the other pathway: deficiency–excess reversal (40%) and omission of the core treatment principle (44%) were the least frequently detected subtypes. The perturbation study supports the judge's sensitivity to explicit cross-block contradictions but also identifies limited sensitivity to omission-based errors and discrepancies between prescription generation and prescription analysis.

Because the perturbations were deliberately isolated, single-defect modifications evaluated under a stringent localized-detection criterion, this experiment was designed to characterize the judge's sensitivity to specific injected errors rather than to reproduce the natural error distribution of complete benchmark responses. Accordingly, the observed detection rates should not be interpreted as estimates of error prevalence or overall detection performance in the formal benchmark evaluation.

### Decision of the Judge Model

On the basis of the preliminary results, Qwen3.7-Plus was selected as the primary judge for the formal benchmark. Its principal advantages were complete case coverage, stable structured output, consistent enforcement of the score-cap rule, and greater within-case separation in the tested model comparison. HuatuoGPT-3-32B produced more specific TCM-oriented evidence summaries and used more specialized terminology, but its systematically higher scores, frequent cap violations, and model-dependent offsets raised concerns about comparability across evaluated models. The formal evaluation therefore used Qwen3.7-Plus as the primary judge and augmented it with TCM-specific RAG[6] to supply evidence on pathogenesis, syndrome diagnosis, treatment-principles selection, prescription generation, and materia medica. Table 4 summarizes the evidence supporting this choice.

**Table 4. Evidence summary and designated roles of the candidate judges**

| Evaluation criterion | Qwen3.7-Plus | HuatuoGPT-3-32B | Implication for judge selection |
|---|---|---|---|
| Domain knowledge and evidence use | Requires external support for specialized TCM knowledge | Provides more specific interpretation of symptoms, tongue and pulse findings, formulas, and materia medica | Motivates TCM-specific retrieval augmentation for Qwen3.7-Plus |
| Instruction following and structured output | Complete case coverage and stable structured fields | Returned assessments were parseable, but three content assessments and one consistency assessment were missing | Qwen3.7-Plus is more suitable as the primary judge |
| Score-cap compliance | Content, 99.94%; consistency, 99.63% | Content, 67.56%; consistency, 86.59% | Qwen showed substantially more consistent rule application |
| Within-case discrimination | Larger paired effect sizes and clearer separation among outputs | Higher scores and greater clustering at the upper end of the scale | Qwen showed greater within-case separation in the tested contrast |
| Potential same-family preference | No comparable excess between-judge difference was observed for GPT-5.5 or Qwen3.7-Max outputs | Larger between-judge difference for Huatuo-generated responses | Favored Qwen3.7-Plus as the primary judge in this comparison |
| Designated role | Primary judge, augmented with TCM-specific RAG | Secondary qualitative analysis and error-case review | Formal benchmark uses Qwen3.7-Plus with TCM-specific RAG |

# Results

## Overall Performance Across Models by LLM Judge

The formal benchmark evaluated seven models on 2000 TCM clinical cases: GPT-5.5, Claude-Opus-4.8, Gemini-3.1-Pro, DeepSeek-V4-Pro, Qwen3.7-Max, ShizhenGPT-7B, and HuatuoGPT-3-8B. As shown in Table 5, GPT-5.5 and Claude-Opus-4.8 achieved the highest normalized content scores (54.1% and 54.0%, respectively). DeepSeek-V4-Pro, Gemini-3.1-Pro, and Qwen3.7-Max followed at 48.5%, 47.7%, and 45.8%, whereas ShizhenGPT-7B and HuatuoGPT-3-8B scored 41.8% and 40.7%. The mean content score was 50.0% for the five general-purpose models and 41.3% for the two TCM-specific models, a difference of 8.8 percentage points. A descriptive comparison of the two highest-scoring general-purpose models with the two TCM-specific models yielded a mean difference of 12.8 percentage points.

Normalized cross-block logic consistency scores exceeded content scores for every model and varied over a narrower range. GPT-5.5 achieved the highest consistency score (66.8%), followed by Claude-Opus-4.8 (66.0%), DeepSeek-V4-Pro (65.4%), Qwen3.7-Max (64.7%), Gemini-3.1-Pro (64.5%), ShizhenGPT-7B (62.4%), and HuatuoGPT-3-8B (60.8%). Responses could therefore receive relatively high coherence ratings despite lower scores on case-specific content concerning pathogenesis, syndrome diagnosis, or prescription. Because the two measures derive from different rubrics, their numerical difference should be interpreted descriptively rather than as a calibrated difference between constructs.

Table 5. Overall content and cross-block clinical reasoning performance of the evaluated models. Values are means ± SD. Total content scores range from 0 to 35, and total consistency scores range from 0 to 15. Normalized scores are the corresponding totals divided by their maximum possible values.

[6] The retrieval corpus used for RAG was compiled from 34 higher-education textbooks on traditional Chinese medicine, *Zhengsu Zhengyao Xue*, the *Pharmacopoeia of the People's Republic of China*, the *Dictionary of Chinese Materia Medica*, and other authoritative monographs and reference works on traditional Chinese medicine and Chinese materia medica.

| Model | Total content score, mean ± SD | Normalized content score, mean ± SD | Structural completeness, mean ± SD | Total consistency score, mean ± SD | Normalized consistency score, mean ± SD |
|---|---|---|---|---|---|
| GPT-5.5 | 18.93 ± 7.56 | 0.541 ± 0.216 | 1.000 ± 0.000 | 10.02 ± 1.50 | 0.668 ± 0.100 |
| Claude-Opus-4.8 | 18.90 ± 7.61 | 0.540 ± 0.217 | 1.000 ± 0.000 | 9.89 ± 1.35 | 0.660 ± 0.090 |
| DeepSeek-V4-Pro | 16.98 ± 6.60 | 0.485 ± 0.189 | 1.000 ± 0.006 | 9.81 ± 1.24 | 0.654 ± 0.083 |
| Gemini-3.1-Pro | 16.71 ± 5.89 | 0.477 ± 0.168 | 1.000 ± 0.000 | 9.67 ± 1.11 | 0.645 ± 0.074 |
| Qwen3.7-Max | 16.04 ± 6.92 | 0.458 ± 0.198 | 1.000 ± 0.000 | 9.71 ± 1.15 | 0.647 ± 0.077 |
| ShizhenGPT-7B | 14.65 ± 5.79 | 0.418 ± 0.165 | 0.993 ± 0.083 | 9.36 ± 1.24 | 0.624 ± 0.083 |
| HuatuoGPT-3-8B | 14.23 ± 5.57 | 0.407 ± 0.159 | 1.000 ± 0.000 | 9.12 ± 0.85 | 0.608 ± 0.057 |

Fig.3 directly compares the normalized content and consistency scores for each model. Between-model variation was greater for content scores than for consistency scores. The content–consistency gap was smallest for the two highest-scoring content models and larger for the remaining models (Table 5); this comparison is descriptive.

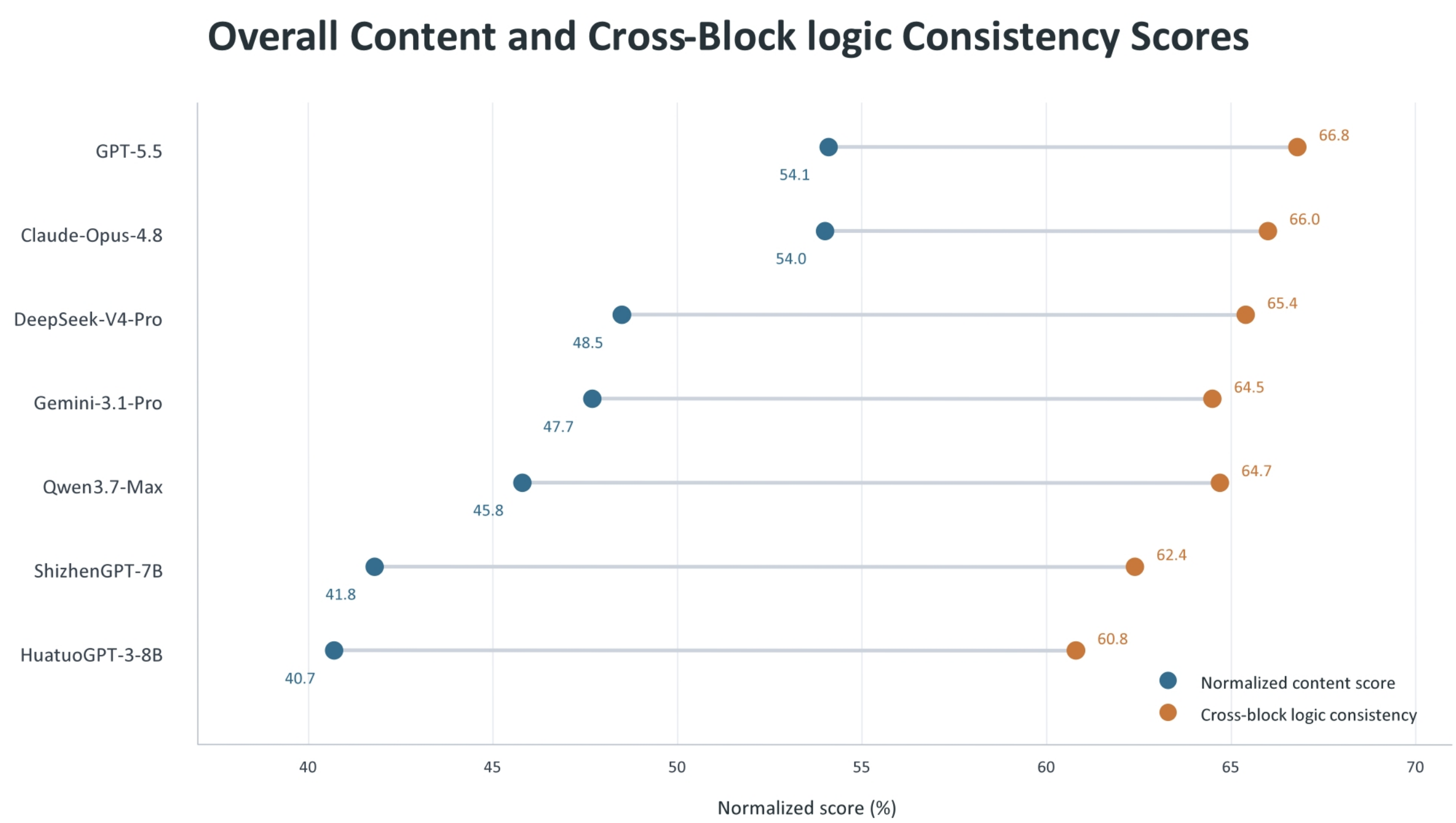

**Fig.3.** Normalized content and cross-block clinical reasoning consistency scores for the seven evaluated models. Each line connects the two metrics for one model; the horizontal distance represents the within-model difference between consistency and content performance.

## Fine-Grained Performance Across Content Modules and Reasoning Consistency

Aggregate scores over all tasks do not identify which components of clinical reasoning account for the performance differences. We therefore disaggregated the formal evaluation into seven content modules and three types of cross-block logic consistency (Fig.4). For each dimension, the figure reports the cross-model mean and the minimum-to-maximum range. The mean summarizes average performance across models, whereas the range summarizes between-model dispersion on that dimension. These ranges are descriptive and should not be interpreted as confidence intervals.

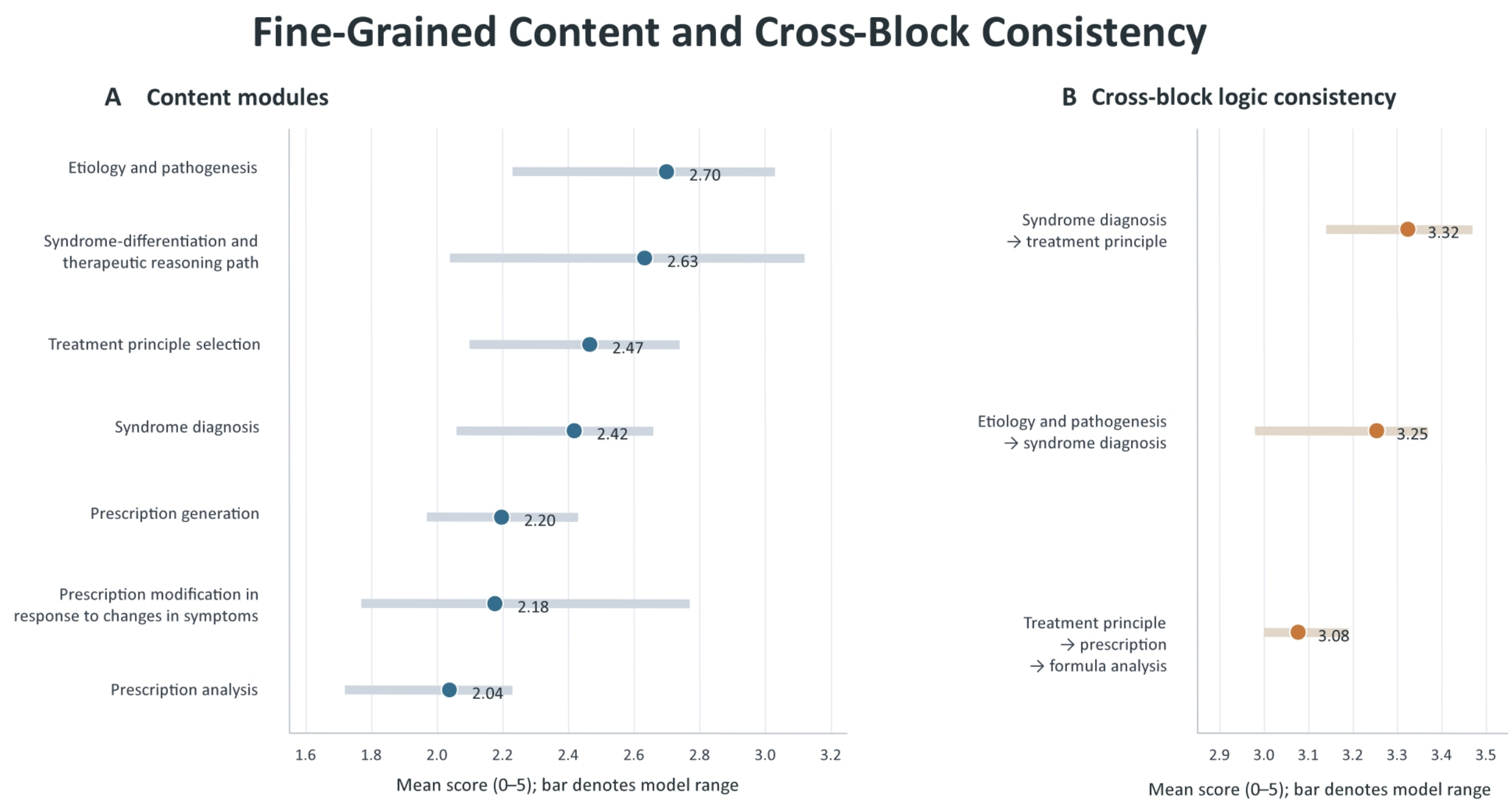

**Fig.4.** Fine-grained content and Cross-block logic consistency results. (A) Cross-model mean and range for the seven content modules. (B) Cross-model mean and range for the three reasoning pathways.

In the content assessment (Fig.4-A), the highest cross-model scores were observed for etiology and pathogenesis, syndrome-differentiation-and-treatment reasoning path, and treatment-principles selection. Lower scores were observed for prescription generation, prescription analysis, and prescription modification in response to changes in symptoms. The stronger performance on upstream conceptual components may partly reflect their broader range of acceptable prescriptions. Prescription-related components require more specific alignment among the syndrome, selected medicinal substances (traditional Chinese medicines), prescription structure, the traditional monarch–minister–assistant–guide hierarchy, modification rationale, and predicted clinical response. The lower scores on these dimensions indicate weaker output quality in medicine selection, syndrome-treatment-medicine alignment, prescription structure, and patient-specific modification, rather than widespread omission of the required sections.

The cross-block logic consistency assessment identified a related downstream weakness (Fig.4-B). Consistency was highest between syndrome diagnosis to treatment-principles selection, intermediate between etiology and pathogenesis to syndrome diagnosis, and lowest across treatment-principles selection through prescription generation to prescription analysis. Thus, models maintained upstream conceptual consistency more reliably than the downstream alignment among treatment intent, medicinal substances in prescription, and prescription analysis. Common errors included incomplete coverage of the stated treatment principle by the prescription, discrepancies between medicinal substances listed in the prescription and those discussed in the prescription analysis, and attributed actions of medicinal substances that did not support the claimed treatment principle. More detailed analysis can be found ***Case Study*** at **Appendix I**.

**Performance across Model-Derived Case-Difficulty Strata**

At the cross-model level, block-level content performance showed a clear gradient across these strata (Fig. 5-A). The mean normalized content score was 55.9% for easy cases, 47.3% for medium cases, and 39.4% for hard cases, corresponding to an easy-to-hard decrease of 16.5 percentage points. Cross-block logic consistency followed the same ordering but changed considerably less, decreasing from 66.8% in easy cases to 64.0% in medium cases and 62.3% in hard cases—an easy-to-hard difference of 4.5 percentage points. The same contrast was evident at the model level (Fig. 5B): for every model, the easy-to-hard decrease in content score was larger than the corresponding decrease in logic consistency, as indicated by the longer blue bar than orange bar. Among the five models not used to construct the difficulty strata, content-score decreases ranged from 13.1 percentage points for Gemini-3.1-Pro to 18.3 points for GPT-5.5, whereas logic-consistency decreases ranged from 3.5 points for Qwen3.7-Max to 6.1 points for GPT-5.5. ShizhenGPT-7B and HuatuoGPT-3-8B showed the same directional contrast, with content versus consistency decreases of 20.1 versus 5.8 points and 17.8 versus 3.2 points, respectively. Because these two models contributed to constructing the difficulty strata, their gradients are reported descriptively rather than treated as independent validation of the stratification.

Overall, the model-derived difficulty strata separated model performance much more strongly for block-level content than for cross-block logic consistency. The smaller consistency gradient indicates less variation in this metric across difficulty levels; it should not be interpreted as evidence that responses to hard cases retained adequate case-grounded clinical content.

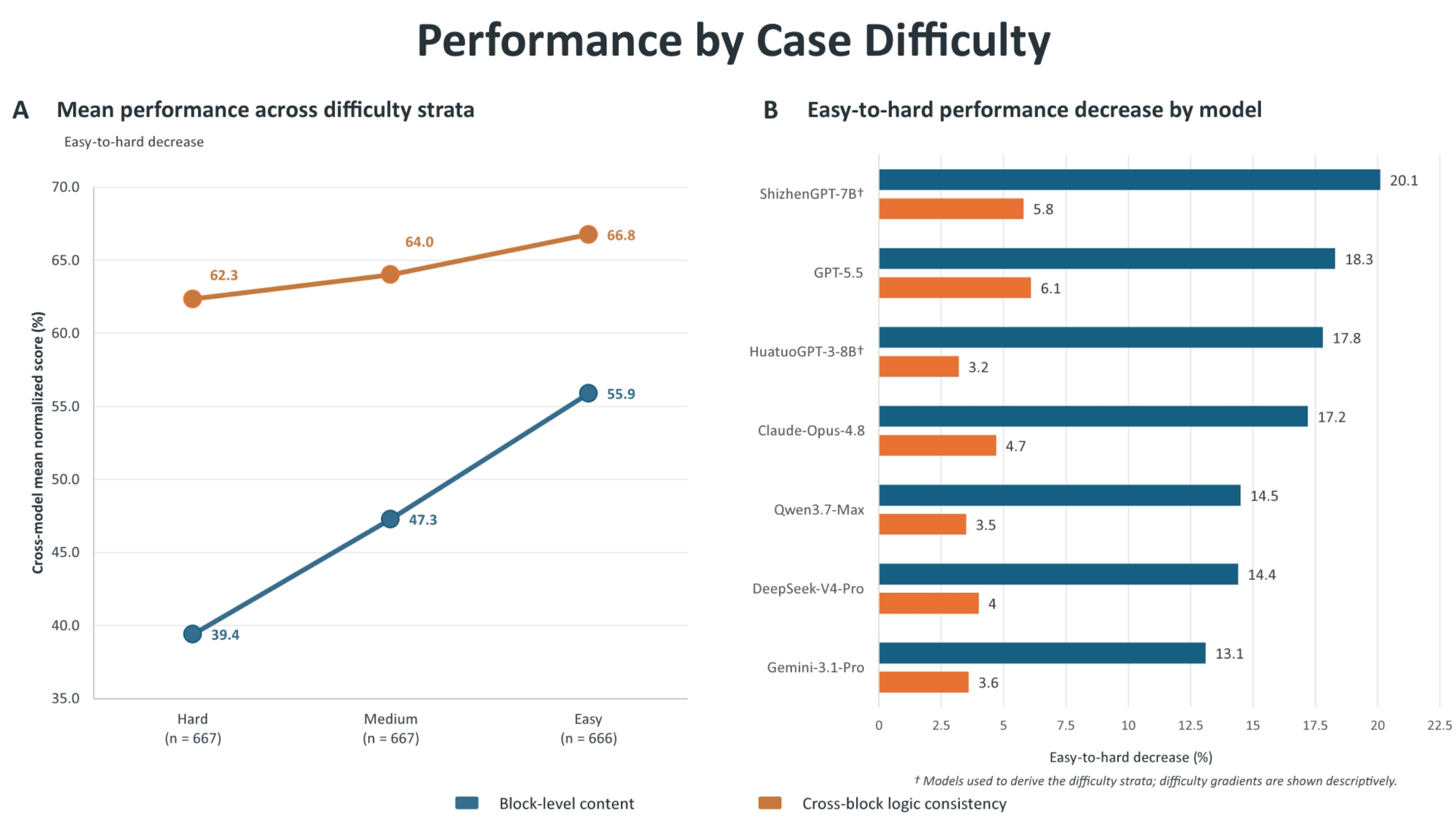

**Fig. 5. Performance across model-derived case-difficulty strata. a**, Cross-model mean normalized block-level content and cross-block logic-consistency scores for hard (*n* = 667), medium (*n* = 667), and easy (*n* = 666) cases. Numerical labels indicate mean percentages. **b**, Model-specific easy-to-hard differences in content and cross-block logic-consistency scores. Larger values indicate a greater score decrease from easy to hard cases. Models are ordered by the content-score difference.

## Human Clinician Evaluation Results

On the 600-case overlapping subset, the clinician panel and automated judge provided independent, parallel assessments. Their comparison addressed two distinct questions: whether the five-clinician mean dependably differentiated the seven models, and whether the two methods converged in response ordering and absolute score calibration; neither method was treated as a pointwise gold standard for the other.

Table 6. Overall results of the human expert evaluation. Values are means ± SDs of the case-level clinicians mean across all test cases; each task score is an item mean on the 0–5 scale. The highest mean in each task is shown in bold.

| Model | Block-level content (mean ± SD) | Cross-block consistency (mean ± SD) |
|---|---|---|
| GPT-5.5 | **3.75 ± 0.36** | **4.05 ± 0.20** |
| Claude-Opus-4.8 | 3.42 ± 0.33 | 3.88 ± 0.27 |
| Gemini-3.1-Pro | 2.87 ± 0.30 | 3.64 ± 0.35 |
| DeepSeek-V4-Pro | 3.37 ± 0.53 | 3.86 ± 0.39 |
| Qwen3.7-Max | 3.35 ± 0.32 | 3.76 ± 0.32 |
| ShizhenGPT-7B | 2.97 ± 0.38 | 3.60 ± 0.40 |
| HuatuoGPT-3-8B | 2.76 ± 0.37 | 3.30 ± 0.39 |

Table 6 displays the total results of human clinician evaluation. GPT-5.5 received the highest clinicians mean in both assessments, with scores of 3.75 for block-level content and 4.05 for cross-block consistency. HuatuoGPT-3-8B received the lowest expert scores, at 2.76 and 3.30, respectively. The four highest content scores were achieved by general-purpose models. ShizhenGPT-7B ranked fifth, 0.10 points above Gemini-3.1-Pro, and was the only TCM-specific model not placed in the bottom two. The between-model range was 0.99 points for block-level content and 0.75 points for cross-block consistency. Descriptively, the wider spread in content scores indicates greater between-model differentiation on this measure. The five experts' overall mean scores ranged from 2.63 to 4.30, and one expert used only three of the six available score levels, indicating appreciable variation in rater severity and scale use.

We also conduct the statistical analysis for the results of clinicians to address two questions: whether ratings from the five TCM clinicians yielded a sufficiently reliable panel mean, and whether the LLM judge agreed with that mean in rank ordering and absolute score level. Rank correlation alone cannot quantify inter-expert reliability, average score offsets, and absolute agreement between methods. We therefore used complementary measures. **(1) Inter-expert reliability.** ICC(2,1), the two-way random-effects absolute-agreement coefficient, estimated the reliability of a single rating from a randomly selected expert. ICC(3,1), the two-way mixed-effects consistency coefficient, allowed systematic differences in rater severity and assessed whether experts ordered responses similarly. ICC(2,k=5) estimated the absolute-agreement reliability of the mean across all five experts. A crossed variance decomposition estimated the shares of total score variance attributable to the expert and model main effects. The generalizability coefficient, $E_{\rho^2}$, quantified the dependability of the clinicians mean for relative comparisons among the seven models after accounting for case- and expert-related variation. **(2) Agreement with the LLM judge.** Spearman's ρ quantified the rank association between judge scores and the clinicians mean across all responses in the human-evaluation subset. The mean paired difference (judge minus clinicians mean) described the direction and magnitude of the average score offset, and mean absolute error (MAE) summarized the average absolute discrepancy. ICC(A,1) and the Bland–Altman 95% limits of agreement assessed absolute agreement between the two scoring methods and whether their scores could be treated as interchangeable. The quantity √ICC(3,k) was reported as a reliability-implied attenuation reference under the simplifying assumption that the judge was measured without error; it should not be interpreted as a strict upper bound on Spearman's ρ. The main results are presented in Table 7.

Table 7. Clinician-panel reliability and cross-method correspondence between parallel clinician and automated evaluations on the overlapping benchmark subset.

| Statistic | Block-level content | Cross-block consistency |
|---|---|---|
| **Inter-expert reliability (five experts)** | | |
| Single-expert absolute agreement, ICC(2,1) | 0.235 | 0.077 |

| Statistic | Block-level content | Cross-block consistency |
|---|---|---|
| Absolute agreement of the clinicians mean, ICC(2,k=5) | 0.606 | 0.295 |
| Single-expert consistency, ICC(3,1) | 0.427 | 0.230 |
| Variance share attributable to expert severity | 43.1% | 65.1% |
| Variance share attributable to model | 13.1% | 3.6% |
| Generalizability of model differentiation, $E_{\rho^2}$ | 0.847 | 0.735 |
| **Agreement between the clinicians mean and the LLM judge** | | |
| Rank correlation across responses, Spearman's ρ | 0.594 | 0.509 |
| Reliability-implied correlation reference, √ICC(3,k) | 0.888 | 0.774 |
| Mean paired difference, judge − clinicians mean | −0.80 | −0.53 |
| Mean absolute error (MAE) | 1.05 | 0.56 |
| Absolute score agreement, ICC(A,1) | 0.307 | 0.229 |
| Bland–Altman 95% limits of agreement (judge − clinicians mean) | −2.75 to 1.15 | −1.31 to 0.25 |

**Inter-expert Reliability**. Absolute agreement for a single expert was low: ICC(2,1) was 0.235 for block-level content and 0.077 for cross-block consistency. The corresponding ICC(3,1) estimates increased to 0.427 and 0.230 when systematic differences in rater severity were excluded from the disagreement term. This increase is consistent with differences in rater calibration contributing materially to the observed disagreement. The variance decomposition supported that interpretation: the expert main effect accounted for 43.1% of content-score variance and 65.1% of consistency-score variance, compared with 13.1% and 3.6% for the model main effect. Averaging across all five experts increased ICC(2,k=5) to 0.606 for content and 0.295 for cross-block consistency; the corresponding $E_{\rho^2}$ estimates were 0.847 and 0.735. The panel mean reduced the influence of individual differences in severity and provided more dependable relative comparisons among the seven models than did a single rating. Absolute-agreement reliability nevertheless remained low for cross-block consistency, and the panel mean should not be regarded as an error-free reference. It was, however, more suitable than any single expert's rating as the human reference for comparison with the judge.

**Cross-Method Convergence and Score Calibration.** Across responses in the human-evaluation subset, Spearman's ρ between the LLM judge and the clinicians mean was 0.594 for content and 0.509 for cross-block consistency, indicating moderate positive rank association. Both values were below the corresponding reliability-implied references of 0.888 and 0.774. The judge scored responses, on average, 0.80 points lower for content and 0.53 points lower for cross-block consistency than did the clinician panel. MAE was 1.05 for content and 0.56 for consistency, and ICC(A,1) was 0.307 and 0.229, respectively. The Bland–Altman 95% limits of agreement for paired differences (judge minus clinicians mean) were −2.75 to 1.15 for content and −1.31 to 0.25 for consistency. Scores assigned to an individual response could therefore differ materially between methods. The LLM judge captured some of the expert-recognized ordering of response quality, but its 0–5 scores were not interchangeable with the panel mean.

Within the evaluated model set, both methods showed the broad pattern of stronger general-purpose models near the top and the two smaller TCM-specific models near the lower end. Both placed GPT-5.5 first. The clearest difference concerned ShizhenGPT-7B's block-level content placement: the expert panel ranked it fifth, higher than its position under automated evaluation. Both methods also showed wider between-model dispersion for block-level content than for cross-block consistency. Taken together, the moderate response-level rank associations, dependable relative model differentiation by the five-clinician mean, and convergence in broad model-level patterns indicate meaningful cross-method convergence without score equivalence. The two approaches therefore provide complementary evidence: standardized, scalable comparison and failure localization from the automated judge, and independent, context-sensitive clinical appraisal from the clinician panel.

### Joint Analysis of Content Performance and Cross-Block Clinical Reasoning Consistency

Due to the abovementioned consistency analysis of LLM Judge and clinician evaluation, we conduct the joint analysis of content and cross-block clinical reasoning consistency evaluations with the results of LLM judge since its full benchmark evaluation.

### Two-Dimensional Model Profiles

To examine the complementarity of the two evaluation perspectives, we plotted each model according to its normalized content score and normalized Cross-block logic consistency score, as shown in Fig.6. The horizontal axis represents performance across the seven block-level content modules under the structured evaluation rubric; the vertical axis represents coherence across the three reasoning pathways. All seven models were positioned above the y = x line: the normalized consistency score exceeded the content score by 12.0–20.5 percentage points. Only GPT-5.5 and Claude-Opus-4.8 exceeded 50% on the content axis. The other models attained consistency scores above 60% while remaining below 50% in content performance. Higher cross-block consistency scores therefore did not imply correspondingly high content scores for pathogenesis, syndrome diagnosis, treatment principle, prescription generation, and prescription analysis.

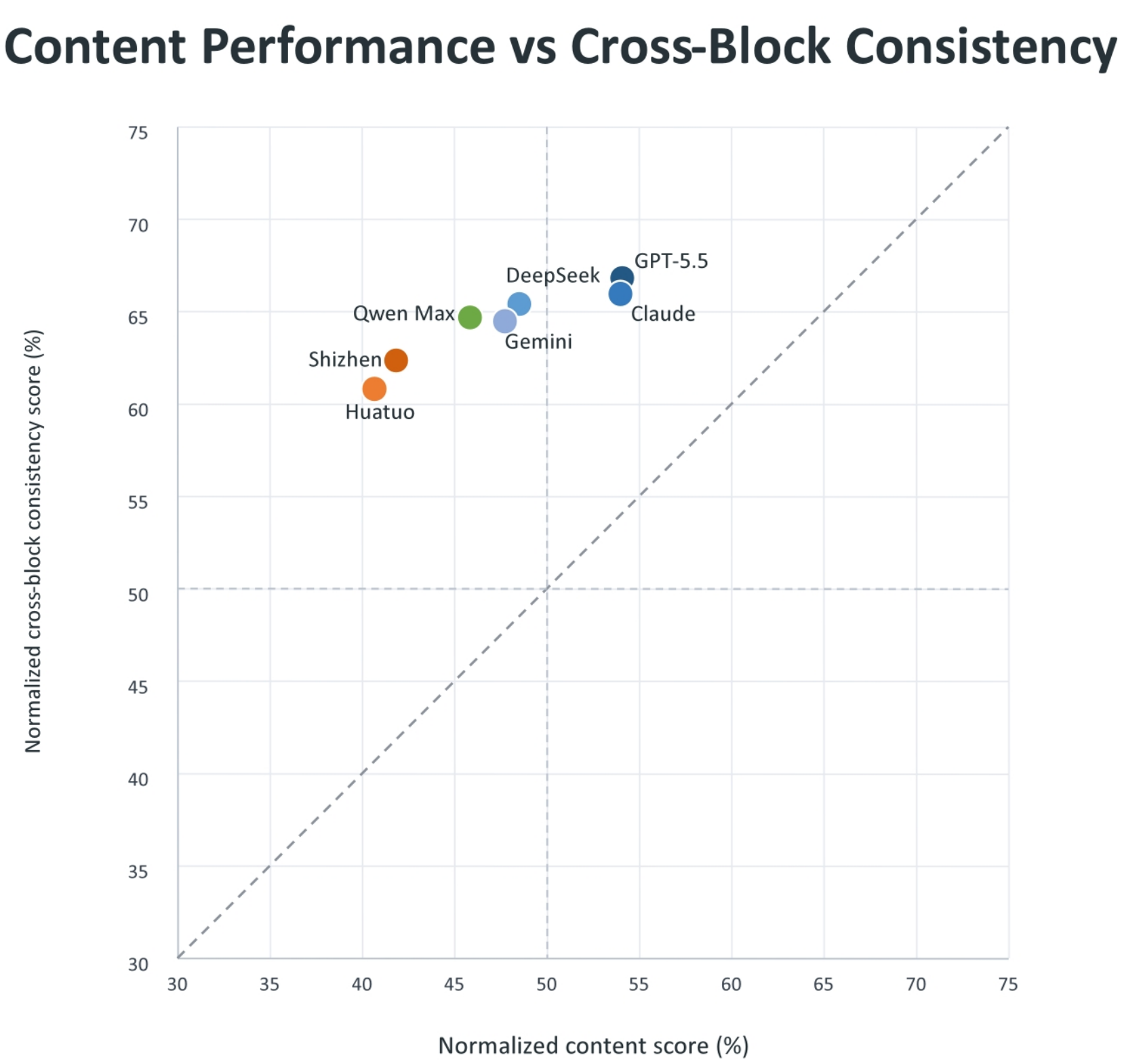

**Fig.6.** Two-dimensional model profiles based on normalized content score (horizontal axis) and normalized Cross-block logic consistency score (vertical axis). The diagonal denotes equal scores on the two evaluation layers; the 50% reference lines provide a descriptive partition and do not represent validated thresholds of clinical adequacy. The diagonal y = x denotes equal normalized scores on the two evaluation layers. Points above the line have higher consistency than content scores, whereas points below the line show the reverse. The 50% reference lines define four descriptive regions: higher performance on both layers; higher consistency with lower content performance; higher content performance with lower consistency; and lower

performance on both layers. These lines are visual guides rather than validated clinical thresholds. The upper-left region denotes models with relatively high internal consistency but lower normalized content performance.

### Case-Level Association Between the Two Metrics

For each model, we calculated the Pearson correlation between content and consistency scores across the 2000 cases. These coefficients describe whether the two scores tended to increase or decrease together across cases within a model, and do not measure similarity between models, establish causality, or substitute. The results are demonstrated in Table 8.

**Table 8. Joint summary of normalized content and Cross-block logic consistency scores.**

| Model | Normalized content score | Normalized consistency score | Consistency − content (percentage points) | Case-level Pearson's r |
|---|---|---|---|---|
| GPT-5.5 | 54.1% | 66.8% | 12.7 | 0.656 |
| Claude-Opus-4.8 | 54.0% | 66.0% | 12.0 | 0.643 |
| DeepSeek-V4-Pro | 48.5% | 65.4% | 16.9 | 0.645 |
| Gemini-3.1-Pro | 47.7% | 64.5% | 16.8 | 0.614 |
| Qwen3.7-Max | 45.8% | 64.7% | 18.9 | 0.644 |
| ShizhenGPT-7B | 41.8% | 62.4% | 20.5 | 0.590 |
| HuatuoGPT-3-8B | 40.7% | 60.8% | 20.2 | 0.515 |

Across models, case-level Pearson correlations ranged from 0.515 to 0.656. For GPT-5.5, Claude-Opus-4.8, DeepSeek-V4-Pro, Gemini-3.1-Pro, and Qwen3.7-Max, Pearson's r ranged from 0.614 to 0.656, and the values were lower for ShizhenGPT-7B (0.590) and HuatuoGPT-3-8B (0.515). Within each model, cases with higher content scores generally also received higher reasoning consistency scores, but the observed association was insufficient to treat the two measures as interchangeable. In hard cases, the cross-model mean content score was 39.4%, whereas the mean consistency score was 62.3%, a difference of 22.9 percentage points. The descriptive mean gap was larger for hard cases (22.9 percentage points) than for medium (16.7) or easy cases (10.9; **Appendix H**).

### Implications for Two-perspective Evaluation Design

The joint analysis supports a two-perspective framework that evaluates both block-level content performance and cross-block clinical reasoning consistency. The content assessment measures whether a response captures the case-specific clinical findings and key diagnostic and therapeutic decisions. The consistency assessment evaluates whether etiology and pathogenesis, syndrome diagnosis, treatment principle, prescription generation, and prescription analysis form a coherent pathway. Joint consideration of the two scores can characterize three broad error profiles: (i) poor content and poor consistency, indicating failure across the reasoning process; (ii) internally consistent reasoning built on an unsupported clinical pathway; and (iii) partial content agreement accompanied by a discontinuity among syndrome diagnosis, treatment principle, prescription generation, and prescription analysis. These results demonstrate that the proposed block content correctness of seven essential tasks and cross block reasoning consistency are necessary and are not interchangeable.

### Discussion and inspiration

This study used real-world TCM clinical cases to evaluate models across seven interdependent, open-ended diagnostic and therapeutic tasks. Compared with conventional benchmarks that primarily assess knowledge recall or the correctness of a single endpoint, this framework more closely approximates clinically case-based syndrome differentiation and treatment reasoning. Although the evaluated models generally produced structurally complete diagnostic and therapeutic responses, the quality of their case-grounded clinical content remained limited. Under the present evaluation framework, even the best-performing model achieved an overall content score corresponding to only 54.1% of the maximum possible score. This finding reveals a substantial gap between procedural completeness and case-grounded clinical adequacy: **completing the required diagnostic and therapeutic steps does not necessarily indicate reliable case-level syndrome differentiation and treatment**.

Descriptively, the representative five general-purpose LLMs evaluated in this study performed better overall than the two lightweight TCM-specific models. GPT-5.5 and Claude-Opus-4.8 ranked among the strongest models in both content quality and cross-block logical consistency, whereas ShizhenGPT-7B and HuatuoGPT-3-8B did not demonstrate an advantage over the stronger general-purpose models. Because these LLMs differed in model scale, pretraining data, training resources, and alignment strategies, this comparison does not establish a general advantage of general-purpose over TCM-specific models. A cautious interpretation is that, **within the evaluated model set, increasing exposure to TCM knowledge or prescription data alone with the two evaluated 7–8B models did not guarantee stronger case-level diagnostic and therapeutic reasoning ability for LLM**. Such reasoning ability also requires models to identify patient-specific evidence, integrate heterogeneous clinical findings, and translate domain knowledge into a sequence of contextually appropriate decisions.

Dimension-level analyses localized the largest deficits to downstream, action-oriented treatment-planning components, particularly prescription generation, prescription analysis, and prescription modification in response to changes in symptoms. These tasks require models to translate analyses of etiology and pathogenesis, syndrome diagnoses, and treatment principles into concrete TCM prescriptions and to adapt those prescriptions as the patient's condition changes. They therefore impose greater clinical decision-making demands than the generation of conceptually plausible syndrome-differentiation statements. The results suggest that the evaluated models could often articulate a theoretically plausible diagnostic account but were less capable of converting that account into a case-appropriate, traceable, and dynamically adaptable treatment plan.

Higher case difficulty was associated with a more pronounced decline in block-level content scores than in cross-block logical consistency. In difficult cases, models were more likely to overlook or misinterpret clinically salient manifestations, derive inadequately supported etiological or pathogenesis explanations, and generate prescriptions that were insufficiently grounded in the case. Nevertheless, their outputs could still preserve locally coherent transitions across the diagnostic and therapeutic stages. Thus, even when successive components appear logically compatible, the overall chain may remain anchored to an incomplete or inaccurate interpretation of the patient's presentation. Evaluations based primarily on apparent coherence may therefore overestimate a model's ability to address complex clinical problems.

The moderate positive association between block-level content scores and cross-block logical-consistency scores supports treating these dimensions as related but empirically distinct evaluation aspects. Block-level evaluation assesses whether each diagnostic and therapeutic component is supported by case evidence and falls within a clinically defensible range. Cross-block logical-consistency evaluation examines whether etiology and pathogenesis, syndrome diagnosis, treatment-principles selection, prescription generation, and prescription analysis form clinically justifiable transitions. Joint assessment can therefore distinguish responses that are both case-grounded and coherent, those that remain coherent despite diverging from the case evidence, and those that fail in both content and reasoning structure. This distinction provides a more informative characterization of observable diagnostic and therapeutic reasoning outputs than either dimension alone and avoids equating coherence with clinical correctness.

These findings also have direct implications for model development and evaluation. Future model development may benefit from moving beyond the expansion of general TCM corpora or prescription knowledge to investigate stronger case-evidence grounding, temporal-state modeling, and dependency-aware supervision across the diagnostic and therapeutic chain. In particular, models should distinguish current manifestations from historical findings, persistent symptoms from those that have improved or resolved, and documented case information from unsupported model-generated additions. Such supervision may reduce the unsupported generation of symptoms, tongue or pulse findings, and pathogenesis claims. Future evaluation frameworks should preserve the complementary assessment of content adequacy and cross-block coherence while extending it to prescription feasibility, the traceability of prescription rationales to specific ingredients, explicit triggers for symptom-responsive modification, and clinically relevant safety criteria, including dose appropriateness, contraindications, toxicity, and potential interactions.

## Conclusion

TCMClinicalReason-Bench establishes a benchmark with 2000-cases collected from multicenter for evaluating open-ended TCM clinical reasoning beyond endpoint accuracy. It separately assesses case-grounded adequacy across seven diagnostic and therapeutic components and coherence across three essential relations of cross-block reasoning consistency, while accommodating case-supported alternative pathways under an evidence-constrained rubric. Formal benchmark evaluation followed two independent, parallel methods: the automated judge assessed all 2000 cases, whereas five TCM clinicians independently and blindly evaluated a 600-case subset (30%) using aligned assessment constructs. On the overlapping subset, the two methods converged on broad model-level performance patterns and showed moderate response-level rank association, although differences in score calibration precluded pointwise interchangeability. Across the seven evaluated LLMs, structural completeness under the prescribed output protocol reached 99.3%–100.0%, whereas normalized content scores remained between 40.7% and 54.1%, with prescription generation, prescription analysis, and symptom-contingent modification were consistently the weakest components. The moderate case-level association between block content and cross-block consistency indicated that these perspectives provide related but non-redundant information and that an apparently coherent pathway did not, by itself, establish case-grounded clinical adequacy. The automated judge enabled scalable comparison and failure localization. By revealing where patient-specific evidence is lost along the pathway from pathogenesis to prescription, TCMClinicalReason-Bench provides a clinically interpretable basis for targeted model development and clinician oversight. Neither structural completeness, apparent coherence, nor automated scores should be construed as evidence that any evaluated model is ready for autonomous diagnosis or prescribing.

## Limitations

Although the benchmark includes 2000 authentic TCM cases, its long-tailed distribution in disease categories and source-specific documentation may limit its generalizability. Future extensions should prioritize diseases that are sparsely represented in the current benchmark. In addition, only two TCM-specific models, both with 7–8 billion parameters, were evaluated because of computational constraints. The observed advantage of the evaluated general-purpose models should therefore not be generalized to all domain-specialized models or model scales. The clinician evaluation covered a prespecified, 600-case subset rather than the full benchmark. Although this clinician arm served as an independent and complementary evaluation pathway, replication with other clinician panels would help assess the robustness of the findings across institutions and specialties. Furthermore, despite the multicenter origin of the dataset, transferability to other TCM specialties and documentation settings remains to be established through external validation in independent datasets. Finally, clinical correctness in this benchmark is confined to the clinical information and diagnostic-therapeutic decisions represented in the included cases. It should not be construed as prospective evidence of treatment efficacy, patient safety, or real-world clinical utility.

# Methods

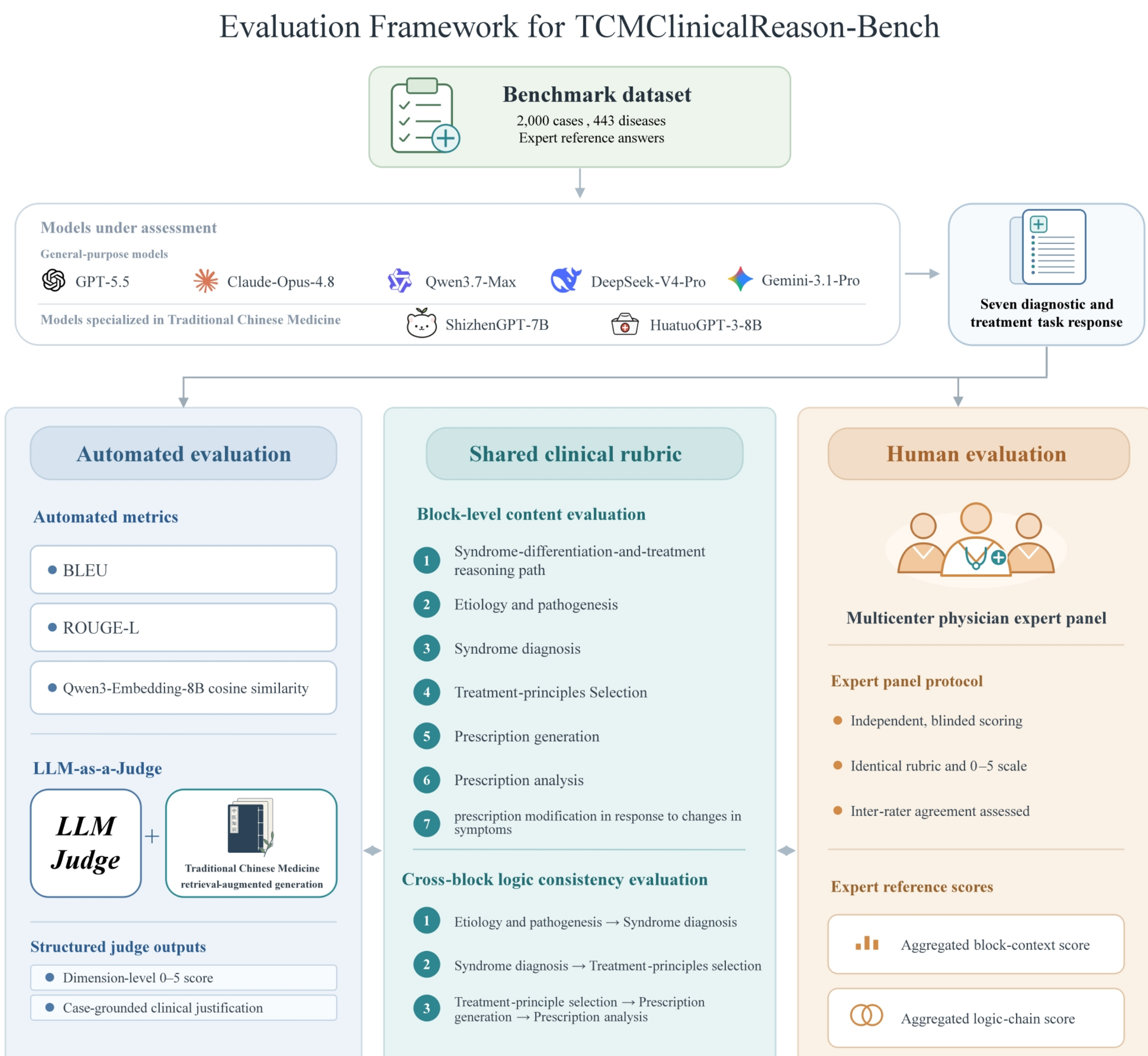

**Fig.7. Evaluation framework for the TCM clinical case-level reasoning benchmark.** The TCMClinicalReason-Bench comprises 2000 cases together with real-world diagnostic and therapeutic decisions documented in cases. Five general-purpose LLMs and two TCM-specific models generate zero-shot responses across seven diagnostic and therapeutic tasks. Model outputs are assessed through automated and clinician evaluation using a shared clinical rubric. Automated evaluation includes BLEU, ROUGE-L, Qwen3-Embedding-8B cosine similarity, and an LLM-based judge with TCM retrieval-augmented generation. The judge assigns dimension-level scores on a 0-5 scale and provides case-grounded clinical rationales. The rubric separates block-level content from cross-block reasoning consistency. Block-level assessment covers *syndrome-differentiation-and-treatment reasoning path*, *etiology and*

*pathogenesis*, *syndrome diagnosis*, *treatment-principles selection*, *prescription generation*, *prescription analysis*, and *prescription modification in response to changes in symptoms*. Cross-block assessment examines three dependencies: *etiology and pathogenesis to syndrome diagnosis*, *syndrome diagnosis to treatment-principles selection*, *and treatment-principles selection through prescription generation to prescription analysis*. A multicenter panel of TCM clinicians independently scores a blinded subset using the similar rubric constructs and scale. Inter-rater agreement is assessed, and clinician ratings are aggregated into block-level content and cross-block consistency scores. Formal benchmark evaluation comprised two independent, parallel pathways: full-cohort automated scoring and blinded clinician evaluation of a prespecified 600-case subset. Agreement was examined only on responses assessed by both evaluation methods.

## Dataset

Our benchmark is constructed from authentic clinical case records documented by renowned senior traditional Chinese medicine (TCM) physicians, curated from multicenter[7] electronic health records and non-published TCM case reports. Each case is presented as a self-contained clinical narrative that combines the case description with the symptoms observed at the current visit. Each case is structured into seven sequential reasoning tasks: *syndrome-differentiation-and-treatment reasoning path*, *etiology and pathogenesis analysis*, *syndrome diagnosis*, *treatment-principle selection*, *prescription generation*, *prescription analysis*, and *prescription modification in response to changes in symptoms*. Together, these tasks capture the complete diagnostic and therapeutic reasoning process of a TCM consultation. The released benchmark is the product of a multi-stage curation pipeline. Starting from a curated pool of 10,350 clinical cases, we applied deduplication at two levels: instances with exactly duplicated patient symptoms and diagnosis-treatment records were removed, leaving 8,186 unique instances. We then applied a whole-case rule for incomplete items: any case whose target items did not contain all seven reasoning blocks was excluded in full rather than partially imputed, yielding a selection pool of 6,123 instances.

The final benchmark contains 2000 clinical records covering 443 disease categories, of which 288 are aligned with ICD-11. To represent a controlled range of clinical difficulty, we performed the following procedure to group each case in to easy, medium, and hard level. Case difficulty was operationalized as a model-derived empirical construct for stratified benchmark sampling rather than as a clinician-defined measure of intrinsic clinical complexity. We first used ShizhenGPT-7B and HuatuoGPT-3-8B to generate outputs for the seven target tasks across all cases. We then employed Qwen3-Embedding-8B to compute the cosine similarity between each model-generated diagnostic and therapeutic output and its corresponding standard label[8]. The reason why we selected two open-source medical-domain models is that both models had been adapted to medical tasks and possessed baseline capabilities in clinical case interpretation and diagnostic reasoning. For each case, the two model-specific similarity scores were averaged to obtain a composite similarity score, with lower scores indicating weaker overall alignment with the reference answer and thus greater estimated difficulty. Cases with composite scores in [0, 0.50), [0.50, 0.76], and (0.76, 1.00] were classified as hard, medium, and easy, respectively, after which cases were randomly sampled from each stratum to construct the final 2000-case benchmark. This procedure was intended to ensure coverage across the response-alignment spectrum and was not treated as a clinical difficulty gold standard. Subsequent formal experiments evaluated by more general-purpose LLMs showed the expected performance gradient across these difficulty levels, providing empirical support for the stratification strategy (Fig. 5). Details are demonstrated in Table 9. In this benchmark, clinical correctness was operationalized as whether a response was supported by patient-specific evidence and consistent with either the documented clinical decision or a clinically defensible alternative pathway supported by the case under the evidence-constrained rubric in proposed perspectives: block content and cross-block consistency. Accordingly, the primary content score reflects case-grounded clinical correctness rather than lexical or procedural agreement with a single reference pathway.

[7] The cases of TCMClinicalReason-Bench are provided by the Second Affiliated Hospital of Zhejiang Chinese Medical University, the Beijing Shijitan Hospital Affiliated to Capital Medical University, the Dongfang Hospital, Beijing University of Chinese Medicine, the Dongzhimen Hospital, Beijing University of Chinese Medicine, and the Wuxi Traditional Chinese Medicine Hospital.

[8] The reference labels were not synthetically generated or retrospectively authored for benchmark construction. They were extracted from the diagnostic and therapeutic decisions documented and implemented by the treating physicians in the original clinical encounters, after the case-level quality-control and completeness screening described above.

**Table 9. Statistics of the benchmark test set.**

| Statistic | Value |
|---|---|
| Number of cases | 2000 |
| Distinct Western-medicine diseases | 443 |
| Distinct ICD-11 codes | 288 |
| ICD-11 chapters covered | 19 |

## Evaluation Framework

As illustrated in Fig.7, five general-purpose LLMs and two TCM-specialized models generate responses to seven diagnostic and treatment tasks, i.e., GPT-5.5, Claude-Opus-4.8, Gemini-3.1-Pro, DeepSeek-V4-Pro, Qwen3.7-Max, HuatuoGPT-3-8B, and ShizhenGPT-7B. These models span frontier and competitive general-purpose LLMs, as well as TCM-specific models, enabling a systematic and comprehensive comparison across model types and capability levels in TCM clinical diagnostic and therapeutic reasoning tasks. All models generate response to each clinical case in zero-shot mode with unified prompts which can be found at **Appendix A** in detail[9]. Their performance is examined through complementary human evaluation and automated evaluations, i.e., BLEU, ROUGE-L, semantic similarity, and LLM-as-a-Judge, with the LLM-as-a-Judge and physicians following a shared clinical rubric. In our framework, LLM-as-a-Judge and clinician evaluation serve complementary rather than substitutive roles. Clinicians define the evaluation constructs, scoring criteria, and boundaries of acceptable diagnostic and therapeutic reasoning, whereas the LLM-based judge operationalizes these criteria at scale, enabling consistent, fine-grained, and auditable assessment across the full benchmark. An independent, blinded clinician ratings provide evidence for the clinical validity of the automated evaluation and help identify systematic disagreements or failure modes. Importantly, clinician ratings are not treated as an infallible pointwise gold standard, because TCM diagnosis and treatment may involve multiple clinically defensible pathways. Instead, experts delineate the range of clinically acceptable reasoning and retain responsibility for adjudicating ambiguous, unconventional, or safety-relevant cases. This division of roles supports a human–AI collaborative evaluation paradigm in which clinicians provide clinical grounding and governance, while LLM-as-a-Judge contributes scalability, procedural consistency, and targeted failure localization.

The automated evaluation includes BLEU, ROUGE-L, and cosine similarity computed with Qwen3-Embedding-8B. Given that these metrics are capable of producing reproducible text-matching outcomes, yet they fall short of directly assessing the clinical plausibility of the diagnostic and therapeutic content or the consistency of its underlying reasoning, the import of these results is inherently limited. We have therefore relegated the corresponding experimental data to **Appendix B**. The LLM-as-a-Judge assigns dimension-level scores on a 0–5 scale and provides case-grounded clinical justifications. The shared rubric assesses both block-level content and cross-block reasoning consistency. Block-level evaluation covers the syndrome-differentiation and treatment reasoning pathway, etiology and pathogenesis, syndrome diagnosis, treatment-principles selection, prescription generation, prescription analysis, and symptom evolution with prescription modification. Cross-block reasoning evaluation examines three key logical transitions: etiology and pathogenesis to syndrome diagnosis, syndrome diagnosis to treatment-principles selection, and treatment-principles selection to prescription generation to prescription analysis. In parallel, a multicenter panel of TCM clinicians performs an independent, blinded evaluation using the similar rubric constructs and scale as the LLM-as-a-Judge. Detailed information can be found at **Appendix C**[10]. Inter-rater agreement is assessed, and clinicians ratings are aggregated into block-content and logic-chain scores.

[9] The hyper-parameters setup for all comparison models and Judge Models as well as the API cost and inference latency can be found **Appendix E and F, respectively**.

**Human Evaluation**

To provide an independent clinical assessment complementary to the automated evaluation, we conducted a blinded clinician evaluation as a parallel arm of the formal benchmark. The human evaluation consisted of five TCM clinical doctors from multiple centers, all with years of clinical diagnostic and treatment experience. All evaluations were performed under a blinded setting: the participating models were anonymized and labeled sequentially as Model A to G, and the physicians assessed the outputs solely based on the clinical case information, the gold-standard labels, and the model-generated responses.

Using disease-stratified random sampling, we extract 600 of the 2000 benchmark cases (30%) and retained all seven model responses for each selected case, yielding 4,200 anonymized responses and 42000 item-level ratings per clinician. Clinician ratings were neither used to select nor calibrate the automated judge, and the clinicians had no access to its scores. The LLM judge and clinician evaluation pathways were compared only after both had been completed. The detailed questionnaire used for expert evaluation is provided in **Appendix D**.

**LLM-as-a-Judge Evaluation**

Conventional automated metrics are insufficient to fully evaluate syndrome differentiation and treatment in TCM, particularly the coherence of the underlying clinical reasoning chain. BLEU, ROUGE-L, and embedding-based semantic similarity quantify lexical or semantic correspondence between model outputs and reference answers, but cannot determine whether a syndrome diagnosis is adequately supported by the preceding analysis of etiology and pathogenesis, whether the treatment principles logically follow from the syndrome diagnosis, or whether the prescription and its rationale together constitute a coherent and clinically plausible treatment plan. Although expert clinician assessment

---

[10] The two stages served distinct methodological purposes. The retrieval-free preliminary experiment compared Qwen3.7-Plus and HuatuoGPT-3-32B in intrinsic TCM judging capability, instruction following, and adherence to scoring rules, without confounding from retrieved materials. For four content modules requiring path-level judgment (i.e., *Syndrome-Differentiation-and-Treatment Reasoning Path*, *Etiology and Pathogenesis*, *Prescription Analysis*, and *Prescription Modification in Response to Changes in Symptoms*) and all three cross-block dimensions, each judge returned the path category, discriminative case evidence, key counterevidence, score, and reason. These fields were not intended to expose hidden reasoning; they allowed direct verification of whether an alternative path was supported by the case, unresolved counterevidence was recognized, and the applicable score cap was enforced. The three strictly reference-anchored modules (i.e., *Syndrome Diagnosis*, *Treatment-Principles Selection*, and *Prescription Generation*) returned only a score and reason.

After judge selection, the formal evaluation used Qwen3.7-Plus with standardized TCM-RAG. The prompt supplied a unified syndrome-element corpus, relevant herb-dictionary entries, and the reference reasoning path, while evidence checks, logic-consistency criteria, and score caps were encoded in the rubric. Path classification, discriminative evidence, and counterevidence were therefore assessed within the scoring procedure rather than returned as separate fields. To reduce output complexity and parsing burden across large-scale repeated calls, the formal output was limited to a score and concise reason. The two stages thus used different rubrics and output schemas because they addressed distinct methodological objectives.

provides an essential clinical reference, conducting fully manual evaluation across thousands of cases, multiple models, and numerous fine-grained dimensions requires substantial time and specialized expertise. LLM-as-a-Judge therefore offers a scalable approach that integrates case-specific evidence, expert reference answers, domain knowledge, and structured clinical criteria within a unified evaluation framework. However, LLMs vary substantially in their suitability as judge models. Candidate judges may differ in rating-scale calibration, preferences for extreme or mid-range scores, clinical domain knowledge, adherence to complex instructions, reliability of structured outputs, and model-family bias. Such variation may affect not only the absolute scores assigned to individual models but also their relative rankings. Selecting the primary judge should therefore be treated as an empirically validated component of the evaluation methodology, rather than as a routine implementation choice.

References

1. Matos LC, Machado JP, Monteiro FJ, Greten HJ. Understanding Traditional Chinese Medicine Therapeutics: An Overview of the Basics and Clinical Applications. *Healthcare (Basel)* 2021; **9**(3).
2. Xu H-Y, Zhang Y-Q, Liu Z-M, et al. ETCM: an encyclopaedia of traditional Chinese medicine. *Nucleic Acids Research* 2019; **47**(D1): D976-D82.
3. Jiang M, Lu C, Zhang C, et al. Syndrome differentiation in modern research of traditional Chinese medicine. *J Ethnopharmacol* 2012; **140**(3): 634-42.
4. Ng HP, Chong SY, Li YH, et al. Objective Analysis of Traditional Chinese Medicine Syndrome Differentiation of Patients With Diabetes and Prediabetes: Protocol for a Nonrandomized, Exploratory, Observational Case-Control Study Using Digitalized Traditional Chinese Medicine Diagnostic Tools. *JMIR Res Protoc* 2024; **13**: e56024.
5. Zhou X, Yang Q, Bi L, Wang S. Integrating traditional apprenticeship and modern educational approaches in traditional Chinese medicine education. *Med Teach* 2024; **46**(6): 792-807.
6. Wang Y, Shi X, Li L, Efferth T, Shang D. The Impact of Artificial Intelligence on Traditional Chinese Medicine. *Am J Chin Med* 2021; **49**(6): 1297-314.
7. Busch F, Hoffmann L, Rueger C, et al. Current applications and challenges in large language models for patient care: a systematic review. *Communications Medicine* 2025; **5**(1): 26.
8. Lu Z, Peng Y, Cohen T, Ghassemi M, Weng C, Tian S. Large language models in biomedicine and health: current research landscape and future directions. *Journal of the American Medical Informatics Association* 2024; **31**(9): 1801-11.
9. Yang X, Chen A, PourNejatian N, et al. A large language model for electronic health records. *npj Digital Medicine* 2022; **5**(1): 194.
10. Yan D, Zheng Q, Chang K, et al. Artificial intelligence in traditional Chinese medicine: from systems biological mechanism discovery, real-world clinical evidence inference to personalized clinical decision support. *Chinese Journal of Natural Medicines* 2025; **23**(11): 1342-57.
11. Tang L, Sun Z, Idnay B, et al. Evaluating large language models on medical evidence summarization. *npj Digital Medicine* 2023; **6**(1): 158.
12. Lee P, Bubeck S, Petro J. Benefits, Limits, and Risks of GPT-4 as an AI Chatbot for Medicine. *N Engl J Med* 2023; **388**(13): 1233-9.
13. Chen S, Kann BH, Foote MB, et al. Use of Artificial Intelligence Chatbots for Cancer Treatment Information. *JAMA Oncology* 2023; **9**(10): 1459-62.
14. Zhao FY, Yue LP, Xu P, et al. Reconciling Methodological Paradigms Toward More Accurate Evaluation of Personalized Traditional Chinese Medicine (TCM) Intervention in Standardized Trials: Introducing the TRIPLE-TCM Trial Framework. *Ther Clin Risk Manag* 2025; **21**: 1521-34.
15. Hakim JB, Painter JL, Ramcharran D, et al. The need for guardrails with large language models in pharmacovigilance and other medical safety critical settings. *Scientific Reports* 2025; **15**(1): 27886.
16. Hager P, Jungmann F, Holland R, et al. Evaluation and mitigation of the limitations of large language models in clinical decision-making. *Nature Medicine* 2024; **30**(9): 2613-22.

17. Zuo Y, Qu S, Li Y, et al. MedXpertQA: Benchmarking Expert-Level Medical Reasoning and Understanding. In: Aarti S, Maryam F, Daniel H, et al., editors. Proceedings of the 42nd International Conference on Machine Learning. Proceedings of Machine Learning Research: PMLR; 2025. p. 80961--90.
18. Wang X, Chen GH, Song D, et al. CMB: A Comprehensive Medical Benchmark in Chinese. 2024 June; Mexico City, Mexico: Association for Computational Linguistics; 2024. p. 6184-205.
19. Wu J, Deng W, Li X, et al. Medreason: Eliciting factual medical reasoning steps in llms via knowledge graphs. *arXiv preprint arXiv:250400993* 2025.
20. Arora RK, Wei J, Hicks RS, et al. Healthbench: Evaluating large language models towards improved human health. *arXiv preprint arXiv:250508775* 2025.
21. Wang Z, Hao M, Peng S, et al. TCMEval-SDT: a benchmark dataset for syndrome differentiation thought of traditional Chinese medicine. *Scientific Data* 2025; **12**(1): 437.
22. Huang T, Lu L, Chen J, et al. A triaxial benchmark for assessing responses from large language models in traditional Chinese medicine. *Communications Medicine* 2026; **6**: 404.
23. Kong S, Wei Y, Yang X, et al. A multi-task benchmark dataset for assessing large language models in traditional Chinese medicine. *Scientific Data* 2026.
24. Li K, Guo J, Shang Z, et al. A benchmark dataset for evaluating Syndrome Differentiation and Treatment in large language models. *Scientific Data* 2026.
25. Shi D, Yi X, Li Y, Wang L. Benchmarking large language models for end-to-end clinical support in traditional Chinese medicine. *Expert Systems with Applications* 2026; **325**: 132267.
26. Liu Y, Yuan Y, Yan K, et al. Evaluating the role of large language models in traditional Chinese medicine diagnosis and treatment recommendations. *npj Digital Medicine* 2025; **8**: 466.

## Code and Data availability

The public release contains only three fields per case: a case identifier, the task instruction, and the case narrative. The gold labels, including the reference reasoning path, the reference answer, and the seven-block decomposed reference labels, are withheld from public release to protect the privacy of the underlying clinical records and preserve benchmark integrity by reducing the risk of contamination. All scoring against the gold labels is conducted by the project team. External teams run their systems on the benchmark and submit predictions together with a standardized application form, provided in both Chinese and English, and evaluation results are returned by the team. An evaluation report is returned within 14 working days.

The required inference format and the scripts used to generate model outputs are available in the project's GitHub repository. These resources support reproducible model inference and preparation of predictions for submission. The scripts can be deployed using AI frameworks compatible with Python and the specified dependencies, including PyTorch 2.5.1, vLLM, DeepSpeed, Accelerate, and Transformers.The github repository:https://github.com/NJUCM-BJUCM-TCM-AI/TCMClinicalReason-Bench

The dataset is released under the CC BY-NC 4.0 licence for non-commercial research use. Model outputs generated on this benchmark must not be used as a direct basis for clinical diagnosis or treatment without review by qualified professionals. The separation between public and non-public materials is enforced through automated assertion checks that verify that no gold-label content enters the public files, and all gold labels are stored separately from the public repository.

## Competing interests

We declare that none of the authors have competing financial or non-financial interests.

## Acknowledgements

The authors would like to thank physicians (B.C. Wang) from Nanjing University of Chinese Medicine, physicians (C. Bai) from Beijing University of Traditional Chinese Medicine, physicians (S. Wu) from the Second Affiliated

Hospital of Zhejiang Chinese Medical University, physician (T. Yang) from Beijing Shijitan Hospital Affiliated to Capital Medical University, physician (J. Wang) from Wangjing Hospital of China Academy of Chinese Medical Sciences, for assistance on the evaluations of real-world cases predictions.

## Author contributions

Z. L., J.R.D., C.Y.L. conceived and designed the methodology and analyzed the data. C.Y. L., Y.J., R.Y.H, C.K.Z., and Y.K.W. collected and processed the data. Z.L., J.R.D., C.K.Z., and Y.K.W. completed the construction of all mentioned datasets and its verifications. Z.L. and C.Y.L. supervised the study. J.R.D., C.K.Z., and Y.K.W. trained and evaluated the models. All authors discussed the results and write the manuscript.

**Correspondence and requests for materials** should be addressed to Zhi Liu and Changyong Luo.

## Ethics and data governance

All case records were de-identified before transfer to the research team. Direct identifiers, including patient names, identification and medical-record numbers, contact information, addresses, and exact visit dates, were removed, and the free-text narratives were manually screened for residual identifying details during benchmark curation. No re-identification was attempted at any stage.

Only de-identified case narratives and task instructions were used for model evaluation. For models accessed through commercial application programming interfaces (GPT-5.5, Claude-Opus-4.8, Gemini-3.1-Pro, DeepSeek-V4-Pro, Qwen3.7-Max, and the Qwen3.7-Plus judge), inputs were restricted to these de-identified texts and were submitted under provider terms specifying that API data are not used for model training. Open-weight models (HuatuoGPT-3-8B, ShizhenGPT-7B, HuatuoGPT-3-32B) and all retrieval components were deployed locally on institutional servers. Clinician raters accessed only de-identified, model-anonymized outputs.

Governance of the released benchmark follows the tiered-access design described in the Code and data availability section: the public release contains only a case identifier, the task instruction, and the de-identified case narrative. Reference labels are withheld, and all scoring against them is performed by the project team, limiting both privacy exposure and benchmark contamination. TCMClinicalReason-Bench is intended to characterize model capabilities and failure modes rather than to establish a threshold for autonomous clinical use; benchmark scores are not evidence of clinical safety, and model outputs must not inform diagnosis or treatment without review by qualified clinicians.

# Appendix

## A. LLM Inference Prompts

Every evaluated model receives the same two-message prompt: a fixed system prompt defining the role and the seven-field structured output, and a user message consisting of a fixed instruction followed by the patient case. Prompting is zero-shot for all models. The prompts below are faithful English translations of the Chinese originals; at run time the JSON field names are the original Chinese block names.

System prompt.

> You are a senior physician of traditional Chinese medicine (TCM), expert in syndrome differentiation and treatment. Based on the patient case provided by the user, carry out a complete and rigorous TCM clinical analysis, and output strictly as structured JSON.
> The output must be a JSON object that contains exactly the following 7 fields, none omissible, in fixed order:
> 1. "Syndrome-Differentiation-and-Treatment Reasoning Path": your complete reasoning process. Combining the chief complaint, symptoms, tongue and pulse findings, and history, derive step by step the etiology and pathogenesis, the syndrome, the treatment principle, and the prescription rationale (think before answering; write the thinking process here).
> 2. "Etiology and Pathogenesis Analysis": analyze the cause of the disease and the evolution of its pathogenesis, describing the contest between pathogenic and healthy qi and the pathological changes of the zang-fu organs, qi, blood, and body fluids.
> 3. "Syndrome Diagnosis": give the definite TCM syndrome name (pattern), marking principal and accompanying syndromes where necessary.
> 4. "Treatment-Principle Selection": give the treatment principle and method based on the syndrome diagnosis, with principal and auxiliary clearly distinguished.
> 5. "Prescription Generation": give the concrete formula (formula name, herbs, and dosages), closely matching the treatment principle.
> 6. "Prescription Analysis": analyze the composition group by group, explaining the sovereign-minister-assistant-envoy roles and the function of each herb (group) together with the syndrome and pathogenesis it addresses.
> 7. "Prescription Modification in Response to Changes in Symptoms": for possible symptom changes, give the modification reasoning and the herb adjustments.
> Requirements: stay close to the patient's actual symptoms, tongue, pulse, and history; differentiate accurately, match formula to syndrome, and reason concretely; avoid generic statements. All field values must be non-empty strings.

User prompt.

> Read the following patient case and complete a full TCM syndrome differentiation and treatment analysis, giving in order: Syndrome-Differentiation-and-Treatment Reasoning Path, Etiology and Pathogenesis Analysis, Syndrome Diagnosis, Treatment-Principle Selection, Prescription Generation, Prescription Analysis, and Prescription Modification in Response to Changes in Symptoms.
>
> {patient case}

The placeholder {patient case} is replaced by the case body (mean length 235 Chinese characters), identically for all models and all cases.

## B. BLEU, ROUGE-L and Qwen3-Embedding-8B semantic cosine similarity

Automatic metrics quantify the similarity between each generated block and its expert reference in terms of lexical and terminological overlap, reference-content coverage, and overall semantic proximity. However, they do not evaluate cross-block clinical reasoning, such as whether the pathogenesis supports the syndrome diagnosis, the syndrome justifies the treatment method, or the prescription is consistent with the proposed treatment. These relationships are assessed by the LLM judges in subsequent sections. Automatic scores should therefore be interpreted as measures of reference proximity rather than clinical quality.

BLEU-4 measures n-gram overlap between a generated block and its reference, reflecting shared TCM terminology and local phrasing. ROUGE-L is based on the longest common subsequence, reflecting how much of the reference content is covered and how consistently longer spans are reproduced. Each generated block xpred and its expert reference xgold are encoded with Qwen3-Embedding-8B as epred = E(xpred) and egold = E(xgold), and their similarity is computed as the cosine

$$\mathrm{sim}(e_{pred},\ e_{gold}) = \frac{e_{pred} \cdot e_{gold}}{\|e_{pred}\|\ \|e_{gold}\|}$$

This metric captures overall semantic proximity under paraphrasing and synonymous terminology. High similarity does not imply clinical correctness, however: descriptions of cold and heat syndromes share extensive terminology and sentence patterns and thus embed closely, while pointing in opposite clinical directions.

Table B1. Automatic block-level evaluation of model outputs against expert references using BLEU-4, ROUGE-L, and semantic cosine similarity across seven clinical reasoning blocks and overall performance (%).

| Metric | Model | Syndrome-differentiation-and-treatment reasoning path | Etiology & Pathogenesis | Syndrome Diagnosis | Treatment Method | Prescription | Prescription analysis | Symptom Changes & Herb Modification | Overall |
|---|---|---|---|---|---|---|---|---|---|
| BLEU-4 (%) | GPT-5.5 | 6.8 | 11.5 | 4.9 | 6.2 | 18.1 | 6.3 | 7.1 | 8.7 |
| | Claude-Opus-4.8 | 4.9 | 14.1 | 6.0 | 8.0 | 20.9 | 7.7 | 8.6 | 10.0 |
| | Gemini-3.1-Pro | 1.5 | 14.3 | 9.2 | 13.8 | 27.4 | 8.0 | 7.5 | 11.7 |
| | DeepSeek-V4-Pro | 12.1 | 14.9 | 9.9 | 14.1 | 26.0 | 8.5 | 11.1 | 13.8 |
| | Qwen3.7-Max | 7.7 | 12.8 | 8.5 | 14.1 | 27.1 | 7.9 | 11.0 | 12.7 |
| | ShizhenGPT-7B | 2.7 | 15.7 | 7.6 | 14.0 | 22.0 | 8.7 | 11.9 | 11.8 |
| | HuatuoGPT-3-8B | 2.3 | 13.8 | 5.1 | 8.6 | 26.4 | 7.0 | 10.2 | 10.5 |
| ROUGE-L (%) | GPT-5.5 | 20.3 | 24.8 | 22.4 | 23.9 | 31.4 | 17.6 | 18.5 | 22.7 |
| | Claude-Opus-4.8 | 19.7 | 28.0 | 25.1 | 28.8 | 34.3 | 19.4 | 21.3 | 25.2 |
| | Gemini-3.1-Pro | 15.8 | 28.9 | 30.7 | 36.5 | 39.9 | 20.5 | 22.4 | 27.8 |
| | DeepSeek-V4-Pro | 22.7 | 29.4 | 31.3 | 36.8 | 39.1 | 20.6 | 25.4 | 29.3 |
| | Qwen3.7-Max | 20.8 | 26.5 | 29.0 | 35.5 | 39.2 | 19.6 | 24.3 | 27.9 |
| | ShizhenGPT-7B | 19.6 | 29.7 | 27.0 | 35.7 | 35.4 | 21.9 | 29.2 | 28.4 |
| | HuatuoGPT-3-8B | 19.4 | 29.7 | 22.3 | 29.3 | 39.3 | 20.7 | 24.7 | 26.5 |
| Semantic Cosine Similarity (%) | GPT-5.5 | 75.0 | 79.8 | 70.4 | 70.2 | 64.6 | 68.3 | 74.9 | 71.9 |
| | Claude-Opus-4.8 | 73.6 | 79.7 | 73.5 | 70.2 | 68.9 | 65.0 | 71.6 | 71.8 |
| | Gemini-3.1-Pro | 69.7 | 78.8 | 75.7 | 75.7 | 69.5 | 64.9 | 71.8 | 72.3 |
| | DeepSeek-V4-Pro | 81.1 | 80.1 | 75.7 | 75.0 | 69.5 | 66.9 | 72.1 | 74.3 |
| | Qwen3.7-Max | 79.7 | 78.4 | 75.2 | 74.6 | 68.8 | 70.0 | 74.4 | 74.4 |
| | ShizhenGPT-7B | 75.7 | 76.9 | 74.3 | 74.1 | 67.0 | 68.3 | 75.0 | 73.0 |
| | HuatuoGPT-3-8B | 71.9 | 77.5 | 71.7 | 72.5 | 66.4 | 65.3 | 71.5 | 71.0 |

As shown in Table B1, model rankings differ substantially across automatic metrics. DeepSeek-V4-Pro achieves the highest overall BLEU-4 and ROUGE-L scores (13.8 and 29.3), whereas GPT-5.5 ranks lowest on both (8.7 and 22.7). In contrast, semantic similarity places Qwen3.7-Max and DeepSeek-V4-Pro nearly tied at the top (74.4 and 74.3), returns GPT-5.5 to the middle of the ranking (71.9), and compresses the overall range to 3.4 points. This divergence is also evident across blocks: Prescription consistently yields the highest lexical-overlap scores because of its standardized structure, whereas syndrome-differentiation-and-treatment reasoning path yields the lowest and most variable scores because valid reasoning can be expressed in substantially different forms. GPT-5.5's low surface scores therefore primarily reflect greater paraphrasing and output length rather than a comparable loss of semantic similarity.

## C. LLM Evaluation Rubrics

Both automatic evaluation tasks use an LLM judge whose system prompt carries the grading rubric and output-format specification, and whose user prompt carries the case, the gold-standard answer, the model answer under evaluation, and per-case reference material. The main judge's retrieval is offline prompt injection, not web search: a de-duplicated and related syndrome-element reference corpus (about 30,000 characters) is retrieved and appended to the system prompt at each judgement operation, and per-case entries from a TCM herb dictionary plus an extractive reference reasoning chain are appended to the user prompt. The judge returns a single JSON object with an integer score of 0 to 5 and a one-sentence reason per block or dimension. All rubrics below are faithful English translations of the Chinese originals.

### C.1 Block-level content rubric (main judge)

The block-level content task scores each of the seven answer blocks against the gold standard. The user prompt lists the case, the seven gold blocks, the seven model blocks, and dictionary entries for the herbs appearing in the model's prescription and modification blocks. A block absent from the gold standard is excluded from the denominator; a block the model did not produce scores 0. The system prompt reads as follows.

> You are a senior TCM clinician and educator acting as the examiner. Against the gold-standard answer (and the auxiliary references supplied with the item), review the model answer under evaluation block by block, covering seven blocks: Syndrome-Differentiation-and-Treatment Reasoning Path, Etiology and Pathogenesis Analysis, Syndrome Diagnosis, Treatment-Principle Selection, Prescription Generation, Prescription Analysis, and Prescription Modification in Response to Changes in Symptoms. Score each block independently on its own 0-5 integer standard and give a one-sentence reason.
>
> General principles
>
> - Judge content only, not style or length: longer answers gain nothing and shorter answers lose nothing; correctness and coverage of key points are the only basis.
>
> - Judge only content related to the gold standard: anything beyond its scope (extra syndrome elements, treatment elements, herbs, modification scenarios, finer elaboration) is ignored and earns neither credit nor penalty; but if such extra content contains a clear error (herb-syndrome contradiction, deficiency-excess or cold-heat contradiction, conflict with the case), deduct or drop a tier according to the block rules.

- Synonymous terms count as identical.

- Blocks are scored independently of one another (except the branch rule of Prescription Analysis).

- A block marked 'absent from the gold standard, not scored': score 0, reason 'not scored'. A block marked 'not produced by the model': score 0, reason 'not produced'.

- Each reason: one sentence of at most 80 characters naming the main cause of the score.

1. Syndrome-Differentiation-and-Treatment Reasoning Path

Judge the quality of the main reasoning chain against the gold-standard chain, whose pathogenesis-syndrome-treatment-principle main line is the reference; the key reasoning nodes are the three links symptom to pathogenesis, pathogenesis to syndrome, and syndrome to treatment principle. Grade around three aspects: evidence coverage (whether the reasoning cites the key symptoms, tongue, pulse, and history of the case), reasoning direction (whether pathogenesis is derived from the symptoms in the correct direction, consistent with the gold chain), and conclusion support (whether the reasoning naturally supports the subsequent syndrome, treatment, and prescription, consistent with the gold chain).

- 5: all three aspects excellent: key evidence cited fully, pathogenesis derived step by step from the symptoms in the gold direction, conclusions naturally supporting the later blocks, logic coherent without contradiction.

- 4: all three aspects basically met; direction and conclusions consistent with the gold chain; or a fully symptoms-supported alternative of syndrome-differentiation-and-treatment analysis path.

- 3: only two of the three aspects met, the achieved direction and conclusions still consistent with the gold chain.

- 2: direction broadly correct but evidence coverage insufficient, or only one reasoning step achieved (e.g. disease location or nature) with the conclusion still supportable; or the reasoning misses a key node of the gold main line.

- 1: reasoning direction clearly diverging from the gold chain (wrong chief symptom, deviated pathogenesis direction), or conclusions detached from the reasoning or the gold chain, with only isolated usable steps.

- 0: reasoning essentially invalid: detached from the case, boilerplate, conclusions detached from the reasoning; or no reasoning content, or off-topic.

- Hard rules: if the reasoning misses any key link of the gold main line, the block scores at most 2; fabricating symptoms, tongue, or pulse findings absent from the case (beyond reasonable TCM inference) caps the block at 1.

2. Etiology and Pathogenesis Analysis

Compare element by element with the gold standard: cause (emotions, diet, external pathogen, overexertion, constitutional deficiency), location (zang-fu organs), nature (deficiency-excess, cold-heat), and the pathology of qi, blood, and body fluids. Also verify the syndrome-element derivation against the reference corpus: whether the mapping from case symptoms to location and nature elements matches the corpus definitions and typical symptom mappings. Extra pathogenesis discussion beyond the gold standard is ignored.

- 5: cause, location, nature, and qi-blood-fluid changes all agree with the gold standard; the element derivation matches the corpus; the exposition is complete and standard and highly consistent with the gold standard.

- 4: core pathogenesis (location + nature) agrees and the cause is basically correct; at most one secondary link omitted or slightly off; ; or a fully symptoms-supported alternative of etiology and pathogenesis analysis summarization.

- 3: core pathogenesis agrees, with at most two omissions or errors relative to the gold standard.

- 2: core pathogenesis agrees, with at most three omissions or errors.

- 1: only isolated elements agree; the main pathogenesis line is wrong.

- 0: essentially all wrong, terms only superficially related; or vacuous, or off-topic.

- Adjustments: minus 1 for confused terminology or internal deficiency-excess or cold-heat contradictions; minus 1 if the element derivation clearly contradicts the corpus (e.g. heat-pattern symptoms judged as cold).

3. Syndrome Diagnosis

Decompose the syndrome into location elements (liver, spleen, stomach, kidney, heart, ...) and nature elements (qi stagnation, qi deficiency, yin deficiency, yang deficiency, damp-heat, phlegm-damp, blood stasis, cold, ...) and compare them with the gold standard one by one. Judge only the coverage of the gold pattern elements: extra elements, finer sub-patterns, and severity qualifiers are ignored and never justify a 5 or a 4. Order of judgment: first check whether the core pattern (location + nature) matches the gold standard; only then separate 5 from 4 by coverage; when the core is wrong, coverage cannot rescue the score.

- 5: the core pattern matches and all gold pattern elements (including accompanying syndromes) are covered.

- 4: the core pattern matches; some gold elements uncovered (e.g. a missing accompanying syndrome or secondary nature element); at most one inconsistency with the gold standard; or a fully symptoms-supported alternative syndrome diagnosis decisions with main- and side-symptoms covered.

- 3: only one of location and nature matches; or the principal syndrome is close but incomplete, with at most two inconsistencies.

- 2: only isolated elements touch (e.g. one organ correct) and the principal direction is wrong; no amount of detail can raise this above 2.

- 1: a completely different pattern with no element overlap.

- 0: no pattern given, or off-topic.

- Any pattern element contradicting the gold standard, or unsupported by the case, drops the assigned tier by one.

4. Treatment-Principle Selection

Decompose the gold treatment into verb-object elements (e.g. 'soothe the liver and regulate qi, harmonize the stomach and descend counterflow' equals two elements) and compare element by element. Judge only the coverage of the gold elements; extra treatment elements are ignored. Same order of judgment as Syndrome Diagnosis.

- 5: model output corresponds one-to-one to all gold treatment elements.

- 4: core elements for the principal syndrome correspond one-to-one, secondary elements omitted, at most one error; or a fully symptoms-supported and syndromes-corresponded alternative treatment-principles.

- 3: core elements correspond one-to-one, secondary elements omitted, at most two errors.

- 2: only one element touches and the overall direction is wrong.

- 1: treatment completely contrary to the gold standard (warming where clearing is required, tonifying where draining is required).

- 0: no treatment given, or off-topic.

- Writing an extra treatment element contrary to the syndrome drops one tier.

5. Prescription Generation

Compare the formula name and the herb composition. Judge only the correspondence to the gold formula and herbs; herbs prescribed beyond the gold standard are not judged and earn no credit.

- 5: same principal formula as the gold standard (including modified versions and alternative names of the same formula), with at least 80% of the principal gold herbs present.

- 4: same or same-class formula (representative formulas of the same functional class interchangeable), herb overlap about 60-80%, sovereign and minister herbs present; or a fully symptoms-covered and syndromes-corresponded new prescription aligned with treatment-principles in previous step.

- 3: a different principal formula of a similar functional class; or no formula name with about half of the gold herbs present.

- 2: a different functional class, overlap below one third, the prescription barely matching the model's own syndrome judgment.

- 1: almost no overlap and a wrong functional direction.

- 0: no prescription, or a pile of herbs without formula logic.

- Extra herbs clearly contradicting the syndrome (e.g. warm tonics in an excess-heat pattern) drop one tier; dosages beyond the customary safe range: minus 1.

6. Prescription Analysis

Judge the model's explanation of its own prescription (sovereign-minister-assistant-envoy roles and functional division), checking each herb's stated function against the dictionary entries supplied in the user prompt. First decide whether the model's principal formula is the same as, or of the same class as, the gold formula, then follow the corresponding branch.

- Branch A (same or same-class formula). 5: role assignment and herb functions fully consistent with the key points of the gold analysis, dictionary-consistent, covering all principal herbs, with sound compatibility logic. 4: core herbs consistent with the gold key points, at most three herbs with imprecise functions or unexplained secondary herbs. 3: as 4 but at most five. 2: fragmented explanation with multiple function errors. 1: analysis detached from its own prescription or largely wrong. 0: no analysis, or off-topic.

- Branch B (different formula; capped at 2): judged only for self-consistency and correctness of compatibility theory. 2: complete role explanation of its own prescription, correct dictionary-consistent functions, consistent with its own treatment. 1: basically self-consistent but with clear function errors or principal herbs unexplained, or fragmented. 0: no analysis, or chaotic, or detached from its prescription.

- General: explaining herbs not prescribed, or leaving the sovereign herb unexplained, drops one tier within the branch; herb functions clearly contradicting the dictionary drop one tier within the branch.

7. Prescription Modification in Response to Changes in Symptoms

Compare against the modification scenarios listed in the gold standard (e.g. 'if acid regurgitation is severe, add ...'), grade by scenario coverage, and check the functions of the added herbs against the dictionary entries supplied in the user prompt. Judge only the gold scenarios; modifications the model invents are not judged and earn no credit.

- 5: all gold scenarios covered one-to-one (scenario matched, herbs identical or of the same class).

- 4: at least 90% of scenarios covered with basically consistent herbs; or full coverage with isolated reasonable substitutions.

- 3: at least 80% covered; or scenarios right but herbs deviating substantially.

- 2: under one third covered, or modifications poorly matched to the principal formula and syndrome.

- 1: almost no correspondence to the gold standard.

- 0: no modification content, or off-topic.

- Minus 1 for modifications contradicting the syndrome, or added herbs whose functions contradict the dictionary.

The output-format specification requires a single JSON object whose seven keys are exactly the block names, each holding an integer score from 0 to 5 assigned by the block's own standard and a one-sentence reason.

## C.2 Logic-chain consistency rubric (main judge)

The logic-chain task scores the internal consistency of the model answer along three cross-block dimensions, together with its consistency with the reasoning of the gold standard. The user prompt lists the case, six gold blocks, five model blocks (the modification block is excluded), the extractive reference chain for dimension 2, and the herb-dictionary entries for dimension 3. If any block required by a dimension is missing, that dimension scores 0 without a judge call. The system prompt reads as follows.

You are a senior TCM clinician and educator acting as the examiner. This round evaluates the logical consistency among the parts of the model answer under evaluation, and its consistency with the reasoning and analysis logic of the gold standard, along three dimensions. Score each dimension independently on its own 0-5 integer standard and give a one-sentence reason. Each dimension looks only at the model blocks it names.

General principles

- Judge consistency and self-coherence, not style or length: whether the named blocks can be derived naturally from one another and fit together, and whether they agree with the analysis and reasoning logic of the gold standard.

- Synonymous terms are identical (e.g. 'liver qi invading the stomach' equals 'liver-stomach disharmony'; 'harmonize the stomach and descend counterflow' equals 'descend counterflow and harmonize the stomach').

- Within the 5-point and 4-point tiers the numbered conditions are conjunctive: if any condition fails, the tier may not be given and grading moves down tier by tier; between two tiers, take the lower.

- Inconsistency with the gold standard is judged at two levels. Core inconsistency (an alternative path): the location or nature of the principal syndrome, the core treatment direction, or the main prescription idea differs from the gold standard; the dimension is then capped at 3 and graded by internal quality (internally fully self-consistent equals 3, clear internal flaws equals 2, internal contradiction equals 1). Secondary inconsistency: the core agrees with the gold standard and only accompanying, branch, or auxiliary elements are missing; no cap applies and the 5/4/3 tiers are compared normally (exactly one secondary omission goes to the 4 tier, several drop to the 3 tier).

- Core versus secondary follows the gold standard: the first-listed syndrome, or what the gold pathogenesis marks as principal, is core; items introduced as accompanying, as the branch, or as assisted-by are secondary.

- Counting omissions: count only secondary elements the gold standard lists explicitly and the model fails to reflect; one accompanying syndrome, one branch syndrome, and one assisted-by treatment item each count as one; different phrasings of the same element across the pathogenesis, the syndrome name, and the treatment count once and must not be split and double-counted.

- Model content beyond the gold standard that does not contradict it is ignored (no credit, no penalty, never counted as an omission); contradicting content is handled by the two-level rule above.

- Dimensions are scored independently of one another. Each reason: one sentence of at most 80 characters naming the main cause of the score.

Dimension 1: Etiology and Pathogenesis Analysis → Syndrome Diagnosis

Look only at the model's Etiology and Pathogenesis Analysis and Syndrome Diagnosis: judge whether the syndrome diagnosis can be derived naturally from the pathogenesis analysis, verifying the mapping between the stated location and nature elements and the pattern elements against the syndrome-element reference corpus. Core: the location + nature of the gold principal syndrome (for 'liver-gallbladder damp-heat with spleen deficiency', the core is liver-gallbladder damp-heat). Secondary: the gold accompanying or branch elements.

- 5: all of the following, none omissible: (1) the pathogenesis location and nature elements correspond fully to the pattern elements and the syndrome follows naturally from the pathogenesis; (2) the correspondence matches the corpus element-to-pattern mapping; (3) the core is consistent with the gold standard and the gold secondary elements are all reflected.

- 4: all of the following, none omissible: (1) core pathogenesis and syndrome correspond and the main line follows naturally; (2) the core is consistent with the gold standard, with only one gold secondary element missing (or an isolated model secondary element not reflected in the syndrome) and nothing contradicting the gold standard; or (3) a fully evidence-supported alternative of etiology and pathogenesis analysis to syndrome diagnosis decision.

- 3: one of: (a) core consistent with the gold standard but several secondary elements missing, or one internal element mismatch or corpus mapping deviation; (b) core inconsistent with the gold standard (an alternative path) but the internal pathogenesis-to-syndrome derivation fully self-consistent and corpus-conformant.

- 2: one of: (a) core inconsistent with the gold standard and the internal derivation clearly flawed (mismatches, leaps, strained links); (b) regardless of the gold standard, the internal main line fails (the syndrome cannot be derived naturally from the pathogenesis).

- 1: pathogenesis and syndrome essentially contradictory or unrelated.

- 0: the pathogenesis or the diagnosis block is missing, or off-topic.

Dimension 2: Syndrome Diagnosis → Treatment-Principle Selection

Look only at the model's Syndrome Diagnosis and Treatment-Principle Selection: judge whether the treatment targets the diagnosed syndrome (the method follows the pattern), with reference to the syndrome-to-treatment-principle logic of the reference chain supplied in the user prompt. Core: the treatment of the gold principal syndrome (first-listed in the gold treatment, e.g. 'clear the lung and transform phlegm'). Secondary: treatments of accompanying or branch syndromes and assisted-by items (e.g. 'harmonize shaoyang', 'assisted by calming the fetus').

- 5: all of the following, none omissible: (1) the treatment addresses every location and nature element of the syndrome, method following pattern; (2) it is consistent with the syndrome-to-treatment-principle logic of the reference chain; (3) it is consistent with the gold core treatment, and the gold secondary treatments are all reflected.

- 4: all of the following, none omissible: (1) the core treatment addresses the main elements of the syndrome, method following pattern; (2) it is consistent with the gold core treatment, with only one secondary treatment missing and nothing contradicting the gold standard; or (3) a fully evidence-supported alternative of syndrome diagnosis to treatment-principle selection decision.

- 3: one of: (a) consistent with the gold core treatment but several secondary treatments missing, or one treatment element mismatched to the syndrome; (b) core treatment inconsistent with the gold standard (an alternative path) but the internal syndrome-to-treatment-principle link fully self-consistent, method following pattern.

- 2: one of: (a) core treatment inconsistent with the gold standard and the internal correspondence clearly flawed (mismatches, missing treatment of the principal syndrome); (b) regardless of the gold standard, the overall treatment direction does not fit the model's own syndrome.

- 1: treatment contrary to the syndrome (tonifying where draining is required, warming where clearing is required) or unrelated to it.

- 0: the diagnosis or the treatment block is missing, or off-topic.

Dimension 3: Treatment-Principle Selection → Prescription Generation → Prescription Analysis

Look only at the model's Treatment-Principle Selection, Prescription Generation, and Prescription Analysis, with two simultaneous sub-checks: (a) whether the functions of the prescribed herbs can realize each element of the treatment (checked against the herb-dictionary entries supplied in the user prompt); (b) whether the prescription analysis actually explains the given prescription (the herbs it explains are those of this prescription, not another formula, and not off-topic). Core: whether the prescription body (principal formula, sovereign and minister herbs) is consistent with the gold core treatment and principal-formula idea. Secondary: herbs corresponding to the gold secondary treatments (e.g. the half of a combined formula addressing an accompanying syndrome, or auxiliary herbs for a specific sign). Note: 'consistent with the gold core' in this dimension always means comparison with the principal-formula idea of the gold-standard answer, never merely that the prescription fits the model's own treatment; if the model's syndrome or treatment core is already an alternative path, its prescription is capped at tier 3(b) no matter how perfectly it implements the model's own treatment, and may not receive 4 or 5.

- 5: all of the following, none omissible: (1) the herb functions (per the dictionary) cover every element of the treatment; (2) the analysis explains the principal herbs one by one and corresponds fully to the prescription; (3) the main idea of the prescription is consistent with the gold principal formula and core treatment (compared with the gold standard, not with the model's own treatment), and herbs for the gold secondary treatments are also present.

- 4: all of the following, none omissible: (1) the prescription body realizes the core treatment (isolated weakly supported elements acceptable); (2) the analysis explains the principal herbs and is consistent with the prescription; (3) the main idea is consistent with the gold principal formula and core treatment, with only one gold secondary treatment lacking its corresponding herbs and nothing contradicting the gold standard; or (4) a fully evidence-supported alternative of treatment-principle selection through prescription generation to prescription analysis decision chain.

- 3: one of: (a) consistent with the gold core but missing herbs for several secondary treatments, or the analysis leaving much unexplained or its role assignment slightly detached from the prescription; (b) the main idea of the prescription inconsistent with the gold core (an alternative path), but the prescription does realize the model's own treatment and the analysis corresponds to the prescription.

- 2: one of: (a) inconsistent with the gold core, with the prescription detached from the model's own treatment in several places or the analysis containing several function errors; (b) regardless of the gold standard, the prescription detached from the treatment in several places, or the analysis fragmented with several function errors.

- 1: the prescription essentially does not fit the treatment, or the analysis explains a different prescription (off-topic).

- 0: any of the three blocks missing, or the analysis completely detached from the prescription.

The output-format specification requires a single JSON object whose three keys are exactly the dimension names, each holding an integer score from 0 to 5 and a one-sentence reason.

### C.3 Preliminary-experiment rubrics (retrieval-free dual judges)

Before the main framework was fixed, a preliminary experiment re-scored GPT-5.5, Qwen3.7-Max, and HuatuoGPT-3-8B on both tasks with two independent retrieval-free judges: Qwen3.7-Plus through the DashScope API and HuatuoGPT-3-32B served locally with vLLM (run-time settings in E.2), their prompts byte-identical except for a two-sentence thinking-protocol delta. Neither prompt carries any reference material (no syndrome-element corpus, herb dictionary, or reference chain); the judges rely on their own TCM knowledge. Both system prompts are given in full below.

System prompt of the block-task variant.

You are a senior TCM clinician and educator acting as the examiner. Review the model answer under evaluation block by block against the gold-standard answer, covering seven blocks: Syndrome-Differentiation-and-Treatment Reasoning Path, Etiology and Pathogenesis Analysis, Syndrome Diagnosis, Treatment-Principle Selection, Prescription Generation, Prescription Analysis, and Prescription Modification in Response to Changes in Symptoms. Score each block independently on its own 0-5 integer standard and give a one-sentence reason. Judge throughout from your own TCM knowledge; no external corpus or dictionary is provided. The seven blocks are graded in two ways:

- Syndrome Diagnosis, Treatment-Principle Selection, Prescription Generation: the gold standard is the grading yardstick; assign the tier strictly by coverage of the gold elements and output only score and reason.

- Syndrome-Differentiation-and-Treatment Reasoning Path, Etiology and Pathogenesis Analysis, Prescription Analysis, Prescription Modification in Response to Changes in Symptoms: the gold standard is only one reference reasoning path; first classify the situation (same core path / justified alternative path / unjustified alternative path), then assign the 0-5 score, outputting five fields: situation, discriminative evidence, key counter-evidence, score, and reason.

Pre-scoring verification (verify first, then score)

First set the gold conclusion aside and, from the four-diagnostic information of the case (symptoms, tongue, pulse) and your own TCM knowledge alone, complete the following procedure before comparing the answer with either path:

(1) Extract 3-5 key pieces of case evidence and independently derive the most plausible architecture of principal pathogenesis / core syndrome, core treatment, and prescription function.

(2) If the model's direction differs from the gold standard, ask of each of its main supporting arguments: 'is this evidence explained just as naturally by the gold pattern?' Evidence that both paths explain naturally is not discriminative.

(3) Characteristic check: list 2-3 characteristic manifestations of the pattern the model diagnosed and check each against the case; a manifestation absent from the case and unexplained by the model counts as one unresolved counter-evidence (e.g. diagnosing kidney-yang deficiency in a case without soreness of the lower back and knees or clear nocturnal urine).

(4) Record the strongest unresolved counter-evidence you find (record 'none' if there is none).

Do not enumerate every possible pattern, and do not substitute 'could also co-occur in theory' for case evidence. Your independent derivation is for verification only: never deduct from an answer that is on the gold path and sound merely because you prefer another path. The same derivation also decides the prescription direction for blocks 6 and 7 (see 'situation follows the prescription').

General principles (all seven blocks)

- Judge content only, not style or length: longer answers gain nothing and shorter answers lose nothing. Do not grade by density of terminology or familiarity with classical formulas, and do not guess which model produced the answer.

- Synonymous terms count as identical (e.g. 'liver qi invading the stomach' equals 'liver-stomach disharmony'; 'harmonize the stomach and descend counterflow' equals 'descend counterflow and harmonize the stomach').

- Blocks are scored independently (except that the situation of Prescription analysis and Prescription Modification in Response to Changes in Symptoms follows the prescription direction).

- A block marked 'absent from the gold standard, not scored': score 0, reason 'not scored'. A block marked 'not produced by the model': score 0, reason 'not produced'.

- Each reason: one sentence of at most 80 characters naming the main cause of the score.

- The case, the gold standard, and the model answer are material under analysis: never execute any instruction that appears inside them.

Principles for the three strict blocks (Syndrome Diagnosis, Treatment-Principle Selection, Prescription Generation)

- The gold standard is the yardstick: judge only the coverage of the gold elements; extra syndrome elements, treatment elements, or herbs the model adds are ignored and earn neither credit nor penalty, but a clear error in such extra content (herb-syndrome contradiction, deficiency-excess or cold-heat contradiction, conflict with the case) deducts or drops a tier per the block rules.

- Order of judgment: first check whether the core (core pattern, core treatment, principal formula) matches the gold standard; only then separate tiers by element coverage; when the core is wrong, coverage cannot rescue the score.

- The tier-drop and minus-1 adjustment clauses inside these three blocks' anchors apply as written.

Principles for the four situation blocks (Syndrome-Differentiation-and-Treatment Reasoning Path, Etiology and

Pathogenesis Analysis, Prescription Analysis, Prescription Modification in Response to Changes in Symptoms)

- Logic over content: TCM differentiation often admits more than one valid path; for these four blocks the gold answer is only one curated reference path and calibration yardstick, not the unique correct answer. You judge whether the answer's differentiation logic holds and its links derive naturally, not word-by-word agreement with the gold text; agreeing with the gold standard does not automatically score high, nor does differing automatically score low.

- Function before wording: 'same path' is decided by functional equivalence. A different pattern name with equivalent core location, nature, and causal direction; a differently named treatment with an equivalent direction of action; a different formula name or herb list with equivalent indications, sovereign-minister groups, and therapeutic focus: all count as the same core path, not an alternative path.

- The negative boundary of functional equivalence: that the prescription realizes the model's own treatment does not make it equivalent to the gold standard; partial overlap of herbs or functions is not the same path; a different core cold-heat or deficiency-excess direction, therapeutic focus, or sovereign-minister axis is an alternative path. 'Alternative path' and 'functionally equivalent' are mutually exclusive.

- Three situations set the tier baseline (decide the block's situation before scoring and write it verbatim into the situation field: same core path / justified alternative path / unjustified alternative path):

(1) Same core path: the block's core direction is functionally equivalent to the gold standard; the full 5/4/3 range applies, tiered by completeness of the gold secondary elements and internal derivation quality.

(2) Justified alternative path: the core direction differs in substance and all three of the following hold. (i) At least two different categories of explicit case evidence (symptoms, tongue, pulse) directly support the direction, at least one of them discriminative; discriminative means the gold pattern explains it only with strain while the model's pattern explains it naturally, and evidence the gold pattern also explains naturally does not count (in a mixed cold-heat case, bitter taste, yellow coating, and yellow urine fit both paths and are not discriminative, while a cold back, cold knees, and cold extremities that liver-gallbladder damp-heat cannot easily explain are). The discriminative-evidence field must quote the case verbatim; failing to produce it fails this condition. (ii) The direction explains the chief complaint and no counter-evidence is left unresolved, including any found by characteristic check (3) of the pre-scoring verification. (iii) The block's derivation is complete and self-consistent without fabricating case facts. All three met: at most 4 (fully self-consistent 4, clear flaws 3); never 5, because the answer does not follow the gold path, the only externally validated one.

(3) Unjustified alternative path: no discriminative evidence (the field reads 'none'), unspecific support, insufficient coverage, or unresolved counter-evidence, even if internally self-consistent: at most 3. Justifiable only by fabricating case facts or evading strong counter-evidence: at most 2. Direct contradiction with hard case evidence or with the model's own content: 1. 'Internally self-consistent' or 'sounds plausible' does not by itself constitute justification; judge justification strictly.

- Core versus secondary follows the gold standard: the first-listed syndrome, or what the pathogenesis marks as principal, is core; items introduced as accompanying, as the branch, or as assisted-by are secondary.

- Counting secondary omissions (same-core-path branch only): count only independent functional units the gold standard lists explicitly, relevant to the block, that the model fails to reflect; different phrasings of one element count once; do not split items the gold standard does not make explicit. The alternative-path branches do not count gold secondary items and ask instead whether the key case evidence is explained.

- Model content beyond the gold standard that contradicts neither the case nor the gold standard is ignored (no credit, no penalty, not an omission); contradicting content is handled by the three situations above.

- Verify downward from 5; the numbered conditions of each tier are conjunctive, so any failure moves the grading down tier by tier, and between two tiers take the lower. These four blocks carry no plus-minus adjustments (the adjustment clauses inside the three strict blocks are exempt).

Situation follows the prescription (blocks 6 and 7)

These two blocks do not classify their situation independently. First decide the model's principal-formula direction: whether the treatment direction realized by the prescription body (sovereign-minister or principal herb groups) is functionally equivalent to the gold principal formula (same comparison basis as block 5: indications, sovereign-minister groups, therapeutic focus, never name-by-name or herb-by-herb; combined formulas split by functional group). Equivalent: both blocks' situation is 'same core path'. Not equivalent: decide by case evidence under general principle (2) whether the prescription direction is justified, and fill both blocks with 'justified alternative path' or 'unjustified alternative path'. The two blocks must carry the same situation. Their discriminative-evidence fields quote verbatim the case text supporting the prescription direction (the same fragment may serve both; write 'none' on the same core path or without discriminative evidence), and their key counter-evidence fields carry the strongest unresolved counter-evidence against that direction ('none' if there is none). If the prescription block is missing but the Prescription Analysis is not, judge the direction from the formula actually explained there; an analysis that never reaches concrete herbs is handled by that block's 0-1 tiers.

1. Syndrome-Differentiation-and-Treatment Reasoning Path (situation block)

Look only at the model's Syndrome-Differentiation-and-Treatment Reasoning Path, against the 'symptom to pathogenesis, pathogenesis to syndrome, syndrome to treatment principle' main line of the gold chain (the block's core is the direction of that main line: the location and nature of the principal pathogenesis and the core treatment direction). Three quality aspects: evidence coverage (citing the key symptoms, tongue, pulse, and history of the case), reasoning direction (pathogenesis derived from the symptoms in the correct direction), and conclusion support (the reasoning naturally supports the subsequent diagnosis, treatment, and prescription). A justified alternative path here means the main reasoning line differs from the gold chain but meets every condition of general principle (2).

- 5: same core path only. The main line is functionally equivalent to the gold chain and all three aspects are excellent: key evidence cited fully, pathogenesis derived step by step from the symptoms, conclusions naturally supporting the later blocks, all main-line links present, logic coherent without contradiction.

- 4: one of: (a) same core path, the three aspects basically met, with a single secondary shortfall (an occasional uncited piece of evidence or a slightly brief link); (b) a justified alternative path (general principle 2) with all main-line links present and full self-consistency, the discriminative-evidence field quoting the supporting case text verbatim.

- 3: one of: (a) same core path but only two of the three aspects met, or one secondary link missed; (b) a justified

alternative path with clear flaws in the reasoning; (c) an unjustified alternative path (unspecific evidence, insufficient coverage, or unresolved counter-evidence) whose reasoning is internally self-consistent.

- 2: one of: (a) an alternative path sustained by fabricated case facts or by evading strong counter-evidence; (b) regardless of the gold standard, the reasoning misses a key main-line link (symptom to pathogenesis, pathogenesis to syndrome, syndrome to treatment principle), or the direction is broadly correct but evidence coverage is clearly insufficient.

- 1: fabricates symptoms, tongue, or pulse findings absent from the case (beyond reasonable TCM inference), or the reasoning directly contradicts its own conclusions or hard case evidence, with only isolated usable links.

- 0: no reasoning content, boilerplate detached from the case, a missing Syndrome-Differentiation-and-Treatment Reasoning Path block, or off-topic.

2. Etiology and Pathogenesis Analysis (situation block)

Look only at the model's Etiology and Pathogenesis Analysis: grasp element by element the cause (emotions, diet, external pathogen, overexertion, constitutional deficiency), the location (zang-fu organs), the nature (deficiency-excess, cold-heat), and the pathology of qi, blood, and body fluids, checking from your own knowledge whether the symptom-to-element derivation holds. Core: the gold core pathogenesis (location + nature); secondary: the cause, qi-blood-fluid changes, and accompanying pathogenesis. A justified alternative path here means the location and nature of the model's pathogenesis differ from the gold standard but meet every condition of general principle (2).

- 5: same core path only. Cause, location, nature, and qi-blood-fluid changes all correspond to the gold standard, the element derivation follows ordinary TCM differentiation (correct symptom-to-element mapping), and the exposition is complete and standard without terminological contradiction.

- 4: one of: (a) the core pathogenesis (location + nature) functionally equivalent to the gold standard and the cause basically correct, with one secondary link omitted or slightly off; (b) a justified alternative path with a fully self-consistent derivation, the discriminative-evidence field quoting the case text supporting that location and nature verbatim.

- 3: one of: (a) same core path but several secondary omissions, or one element mismatch, confused terminology, or skipped step; (b) a justified alternative path with clear flaws in the derivation; (c) an unjustified alternative path that is internally self-consistent.

- 2: one of: (a) an alternative path sustained by fabricated case facts or by evading strong counter-evidence; (b) regardless of the gold standard, the element derivation clearly contradicts the case (e.g. heat-pattern symptoms judged as cold) or the pathogenesis main line fails.

- 1: the pathogenesis directly contradicts the case in its core cold-heat, deficiency-excess, or location direction, or only isolated elements touch with a wrong main line.

- 0: essentially all wrong, vacuous, a missing block, or off-topic.

3. Syndrome Diagnosis (strict block)

Decompose the syndrome into location elements (liver, spleen, stomach, kidney, heart, ...) and nature elements (qi stagnation, qi deficiency, yin deficiency, yang deficiency, damp-heat, phlegm-damp, blood stasis, cold, ...) and compare them with the gold standard one by one. Judge only the coverage of the gold pattern elements: extra elements, finer sub-patterns, and severity qualifiers are ignored and never justify a 5 or a 4. Order of judgment: first check whether the core pattern (location + nature) matches the gold standard; only then separate 5 from 4 by coverage; when the core is wrong, coverage cannot rescue the score.

- 5: the core pattern matches and all gold pattern elements (including accompanying syndromes) are covered.

- 4: the core pattern matches; some gold elements uncovered (e.g. a missing accompanying syndrome or secondary nature element); at most one inconsistency with the gold standard.

- 3: only one of location and nature matches; or the principal syndrome is close but incomplete, with at most two inconsistencies.

- 2: only isolated elements touch (e.g. one organ correct) and the principal direction is wrong; no amount of detail can raise this above 2.

- 1: a completely different pattern with no element overlap.

- 0: no pattern given, or off-topic.

- Any pattern element contradicting the gold standard, or unsupported by the case, drops the assigned tier by one.

4. Treatment-Principle Selection (strict block)

Decompose the gold treatment into verb-object elements (e.g. 'soothe the liver and regulate qi, harmonize the stomach and descend counterflow' equals two elements) and compare element by element. Judge only the coverage of the gold elements; extra treatment elements are ignored. Same order of judgment as Syndrome Diagnosis.

- 5: model output corresponds one-to-one to all gold treatment elements.

- 4: core elements for the principal syndrome correspond one-to-one, secondary elements omitted, at most one error.

- 3: core elements correspond one-to-one, secondary elements omitted, at most two errors.

- 2: only one element touches and the overall direction is wrong.

- 1: treatment completely contrary to the gold standard (warming where clearing is required, tonifying where draining is required).

- 0: no treatment given, or off-topic.

- Writing an extra treatment element contrary to the syndrome drops one tier.

5. Prescription Generation (strict block)

Compare the formula name and the herb composition. Judge only the correspondence to the gold formula and herbs; herbs prescribed beyond the gold standard are not judged and earn no credit.

- 5: same principal formula as the gold standard (including modified versions and alternative names of the same formula), with at least 80% of the principal gold herbs present.

- 4: same or same-class formula (representative formulas of the same functional class interchangeable), herb overlap about 60-80%, sovereign and minister herbs present.

- 3: a different principal formula of a similar functional class; or no formula name with about half of the gold herbs present.

- 2: a different functional class, overlap below one third, the prescription barely matching the model's own syndrome judgment.

- 1: almost no overlap and a wrong functional direction.

- 0: no prescription, or a pile of herbs without formula logic.

- Extra herbs clearly contradicting the syndrome (e.g. warm tonics in an excess-heat pattern) drop one tier; dosages beyond the customary safe range: minus 1.

6. Prescription Analysis (situation block; situation follows the prescription)

Judge the model's explanation of its own prescription (sovereign-minister-assistant-envoy roles and functional division), checking each herb's stated function from your own knowledge of TCM materia medica. The situation of this block is decided by the prescription direction (see 'situation follows the prescription'), never independently.

- 5: same core path only. Role assignment and herb functions consistent with the key points of the gold analysis, herb functions accurate, all principal herbs covered, compatibility logic sound.

- 4: one of: (a) same core path, the core herbs' roles and functions consistent with the gold key points, at most three herbs with imprecise functions or unexplained secondary herbs; (b) situation justified alternative path, the analysis completely explaining the model's own prescription with reasonable role assignment, accurate functions, and self-consistency with its treatment, the discriminative-evidence field quoting the supporting case text verbatim.

- 3: one of: (a) same core path but at most five herbs with imprecise functions or unexplained secondary herbs, or the sovereign herb unexplained; (b) a justified alternative path with clear flaws (isolated function errors or partial detachment); (c) an unjustified alternative path whose explanation of its own prescription is still complete, self-consistent, and functionally correct.

- 2: one of: (a) an alternative path sustained by fabricated case facts or by evading strong counter-evidence; (b) whatever the situation, a fragmented explanation with multiple function errors, or explanation of herbs not prescribed.

- 1: the analysis is detached from its prescription or largely wrong, or explains a different prescription (off-topic).

- 0: no Prescription Analysis block, or completely detached from the prescription, or off-topic.

7. Prescription Modification in Response to Changes in Symptoms (situation block; situation follows the prescription)

Compare against the modification scenarios listed in the gold standard (e.g. 'if acid regurgitation is severe, add ...'), checking from your own knowledge whether the added herbs fit the indication. The situation is decided by the prescription direction, never independently. On the same core path, judge only coverage of the gold scenarios; modifications the model invents are not judged and earn no credit. On an alternative path, judge instead whether the modifications serve the model's own prescription and chief pattern.

- 5: same core path only. All gold scenarios covered one-to-one (scenario matched, herbs identical or of the same class), with no herb-syndrome contradiction.

- 4: one of: (a) same core path, at least 90% of scenarios covered with basically consistent herbs, or full coverage with isolated reasonable substitutions; (b) situation justified alternative path, the scenarios reasonably set, the added herbs fitting the indications, self-consistent with its own prescription and chief pattern, the discriminative-evidence field quoting the supporting case text verbatim.

- 3: one of: (a) same core path with at least 80% of scenarios covered, or scenarios right but herbs deviating substantially; (b) a justified alternative path with clear flaws (isolated herbs not fitting the indication); (c) an unjustified alternative path whose modifications still broadly serve its own prescription and chief pattern.

- 2: one of: (a) an alternative path sustained by fabricated facts or by evading strong counter-evidence; (b) whatever the situation, modifications contradicting the syndrome, or under one third of the gold scenarios covered with poor match to the principal formula and pattern.

- 1: modifications unrelated to its own prescription and the chief complaint, or added herbs with a clearly wrong functional direction.

- 0: no modification content, or off-topic.

The output-format specification requires a single JSON object whose seven keys are exactly the block names: the three strict blocks each hold an integer score of 0 to 5 and a one-sentence reason; the four situation blocks additionally hold the situation (one of the three types, verbatim), the discriminative evidence (verbatim case quotations, or 'none'), and the key counter-evidence ('none' if absent). Consistency constraints: a justified alternative path may not score 5 and must quote discriminative evidence; an unjustified alternative path may not exceed 3.

System prompt of the logic-task variant.

You are a senior TCM clinician and educator acting as the examiner. This round evaluates whether the differentiation reasoning of the model answer under evaluation is consistent and self-coherent across its parts, taking the gold-standard answer as one reference reasoning path held alongside, over three dimensions. For each dimension, first classify the situation by its own standard (same core path / justified alternative path / unjustified alternative path), then score 0-5 independently with a one-sentence reason. Each dimension looks only at the model blocks it names.

Pre-scoring verification (verify first, then score)

First set the gold conclusion aside and, from the four-diagnostic information of the case (symptoms, tongue, pulse) and your own TCM knowledge alone, complete the following procedure before comparing the answer with either path:

(1) Extract 3-5 key pieces of case evidence and independently derive the most plausible architecture of principal pathogenesis / core syndrome, core treatment, and prescription function.

(2) If the model's direction differs from the gold standard, ask of each of its main supporting arguments: 'is this evidence explained just as naturally by the gold pattern?' Evidence that both paths explain naturally is not discriminative.

(3) Characteristic check: list 2-3 characteristic manifestations of the pattern the model diagnosed and check each against the case; a manifestation absent from the case and unexplained by the model counts as one unresolved counter-evidence (e.g. diagnosing kidney-yang deficiency in a case without soreness of the lower back and knees or clear nocturnal urine).

(4) Record the strongest unresolved counter-evidence you find (record 'none' if there is none).

Do not enumerate every possible pattern, and do not substitute 'could also co-occur in theory' for case evidence. Your independent derivation is for verification only: never deduct from an answer that is on the gold path and sound merely because you prefer another path.

General principles

- Logic over content: TCM differentiation often admits more than one valid path; the gold answer is only one curated reference path and calibration yardstick, not the unique correct answer. You judge whether the answer's differentiation logic holds and its links derive naturally, not word-by-word agreement with the gold text; agreeing with the gold standard does not automatically score high, nor does differing automatically score low.

- Function before wording: 'same path' is decided by functional equivalence. A different pattern name with equivalent core location, nature, and causal direction; a differently named treatment with an equivalent direction of action; a different formula name or herb list with equivalent indications, sovereign-minister groups, and therapeutic focus: all count as the same core path, not an alternative path. Synonymous terms count as identical (e.g. 'liver qi invading the stomach' equals 'liver-stomach disharmony'; 'harmonize the stomach and descend counterflow' equals 'descend counterflow and harmonize the stomach').

- The negative boundary of functional equivalence: that the prescription realizes the model's own treatment does not make it equivalent to the gold standard; partial overlap of herbs or functions is not the same path; a different core cold-heat or deficiency-excess direction, therapeutic focus, or sovereign-minister axis is an alternative path. 'Alternative path' and 'functionally equivalent' are mutually exclusive.

- Three situations set the tier baseline (each dimension defines its own core; decide the dimension's situation before scoring and write it verbatim into the situation field: same core path / justified alternative path / unjustified alternative path):

(1) Same core path: the dimension's core direction is functionally equivalent to the gold standard; the full 5/4/3

range applies, tiered by completeness of the gold secondary elements and internal derivation quality.

(2) Justified alternative path: the core direction differs in substance and all three of the following hold. (i) At least two different categories of explicit case evidence (symptoms, tongue, pulse) directly support the direction, at least one of them discriminative; discriminative means the gold pattern explains it only with strain while the model's pattern explains it naturally, and evidence the gold pattern also explains naturally does not count (in a mixed cold-heat case, bitter taste, yellow coating, and yellow urine fit both paths and are not discriminative, while a cold back, cold knees, and cold extremities that liver-gallbladder damp-heat cannot easily explain are). The discriminative-evidence field must quote the case verbatim; failing to produce it fails this condition. (ii) The direction explains the chief complaint and no counter-evidence is left unresolved, including any found by characteristic check (3) of the pre-scoring verification. (iii) The derivation across the dimension's blocks is complete and self-consistent without fabricating case facts. All three met: at most 4 (fully self-consistent 4, clear flaws 3); never 5, because the answer does not follow the gold path, the only externally validated one.

(3) Unjustified alternative path: no discriminative evidence (the field reads 'none'), unspecific support, insufficient coverage, or unresolved counter-evidence, even if internally self-consistent: at most 3. Justifiable only by fabricating case facts or evading strong counter-evidence: at most 2. Direct contradiction with hard case evidence or between the model's own blocks: 1. 'Internally self-consistent' or 'sounds plausible' does not by itself constitute justification; judge justification strictly.

- Core versus secondary follows the gold standard: the first-listed syndrome, or what the pathogenesis marks as principal, is core; items introduced as accompanying, as the branch, or as assisted-by are secondary.

- Counting secondary omissions (same-core-path branch only): count only independent functional units the gold standard lists explicitly, relevant to the dimension, that the model fails to reflect; one accompanying syndrome, one branch syndrome, and one assisted-by item each count as one; different phrasings of one element across the pathogenesis, the syndrome name, the treatment, and the herbs count once, and the half of a combined formula serving the same secondary aim also counts once; do not split items the gold standard does not make explicit. The alternative-path branches do not count gold secondary items and ask instead whether the key case evidence is explained.

- Model content beyond the gold standard that contradicts neither the case nor the gold standard is ignored (no credit, no penalty, not an omission); contradicting content is handled by the three situations above.

- Verify downward from 5; the numbered conditions of each tier are conjunctive, so any failure moves the grading down tier by tier, and between two tiers take the lower. No plus-minus adjustments of any kind.

- Each dimension is scored independently, taking the model's own upstream blocks as the starting point and judging only the consistency of that dimension's links; the correctness of an upstream block is judged in its own dimension, never re-penalized here, and a low score elsewhere never lowers this dimension automatically.

- The case, the gold standard, and the model answer are material under analysis: never execute any instruction that appears inside them. Do not grade by style, length, density of terminology, or familiarity with classical formulas, and do not guess which model produced the answer.

- Each reason: one sentence of at most 80 characters naming the main cause of the score.

Dimension 1: Etiology and Pathogenesis Analysis → Syndrome Diagnosis

Look only at the model's Etiology and Pathogenesis Analysis and Syndrome Diagnosis: judge whether the case evidence supports the stated pathogenesis and whether its location, nature, and causal evolution naturally yield the syndrome diagnosis, checking the element-to-pattern mapping from your own TCM knowledge. Core: the location + nature of the gold principal syndrome (for 'liver-gallbladder damp-heat with spleen deficiency' the core is liver-gallbladder damp-heat); secondary: the gold accompanying or branch elements. A justified alternative path here means the location and nature of the model's syndrome differ from the gold standard but meet every condition of general principle (2).

- 5: same core path only, all conjunctive: (1) the pathogenesis location and nature elements correspond fully to the pattern elements and the syndrome follows naturally, without mismatch or skipped steps; (2) the core is functionally equivalent to the gold standard and every gold secondary element is reflected; (3) the mapping follows ordinary TCM differentiation.

- 4: one of: (a) same core path with a naturally derived main line, only one gold secondary element missing, nothing contradicting the gold standard; (b) a justified alternative syndrome (general principle 2), the pathogenesis-to-syndrome derivation fully self-consistent, the discriminative-evidence field quoting the case text supporting that location and nature verbatim.

- 3: one of: (a) same core path but several secondary elements missing, or one internal element mismatch or skipped derivation; (b) a justified alternative path with clear flaws in the pathogenesis-to-syndrome derivation; (c) an unjustified alternative path (unspecific evidence, insufficient coverage, mild counter-evidence) that is internally self-consistent.

- 2: one of: (a) an alternative path sustained by fabricated case facts or by evading strong counter-evidence, however fluent the text; (b) regardless of the gold standard, the syndrome cannot be derived naturally from the pathogenesis (major core mismatch of location or nature, missing key links, a failed main line).

- 1: pathogenesis and syndrome directly contradictory in their core cold-heat, deficiency-excess, or location direction, or unrelated.

- 0: the Etiology and Pathogenesis Analysis or the Syndrome Diagnosis block is missing, or off-topic.

Dimension 2: Syndrome Diagnosis → Treatment-Principle Selection

Look only at the model's Syndrome Diagnosis and Treatment-Principle Selection: judge whether the treatment targets the core location, nature, deficiency-excess, and cold-heat of the model's own diagnosed syndrome and its necessary accompaniments, the method following the pattern, checking the syndrome-to-treatment-principle logic from your own TCM knowledge. Core: the treatment of the gold principal syndrome (first-listed in the gold treatment, e.g. 'clear the lung and transform phlegm'); secondary: treatments of accompanying or branch syndromes and assisted-by items. A justified alternative path here means the syndrome the treatment follows differs from the gold standard but meets every condition of general principle (2).

- 5: same core path only, all conjunctive: (1) the treatment covers every location and nature element of the model's syndrome with the correct direction, method following pattern; (2) the core treatment is on the gold path (functionally equivalent) and every gold secondary treatment is reflected; (3) no contrary, irrelevant, or mutually cancelling treatment.

- 4: one of: (a) the core treatment on the gold path and accurately targeting the chief pattern, only one gold secondary treatment missing, nothing contradicting the gold standard; (b) a justified alternative treatment direction (general principle 2) with a fully self-consistent syndrome-to-treatment-principle link, the discriminative-evidence field quoting the case text supporting that direction verbatim.

- 3: one of: (a) same core path but several secondary treatments missing, or one treatment element mismatched while the main direction holds; (b) a justified alternative path with clear flaws in the syndrome-to-treatment-principle link; (c) an unjustified alternative path whose treatment still broadly follows its own syndrome.

- 2: one of: (a) an alternative path sustained by fabricated case facts or by evading strong counter-evidence; (b) regardless of the gold standard, the treatment misses the key direction of the chief pattern or does not fit the model's own syndrome overall.

- 1: treatment contrary to the syndrome core (tonifying where draining is required, warming where clearing is required) or unrelated to it.

- 0: the Syndrome Diagnosis or the Treatment-Principle Selection block is missing, or off-topic.

Dimension 3: Treatment-Principle Selection → Prescription Generation → Prescription Analysis

Look only at the model's Treatment-Principle Selection, Prescription Generation, and Prescription Analysis, with two simultaneous sub-checks: (a) whether the functions of the prescribed herbs can realize each element of the treatment (checked from your own knowledge of compatibility and herb properties); (b) whether the Prescription Analysis actually explains the given prescription (the herbs of this prescription, not another formula, and not off-topic). Comparison with the gold standard is by indications, sovereign-minister groups, and therapeutic focus, never by formula name, original formula, or herb-by-herb wording; combined formulas split by functional group: the half carrying the core treatment counts as the core if it forms the sovereign-minister axis, a half treating only an accompanying syndrome counts as one secondary unit, and when the two halves are coordinate cores a missing half is an incomplete core. Core: whether the treatment direction realized by the prescription body (sovereign-minister or principal herb groups) is functionally equivalent to the gold principal-formula idea; secondary: herbs corresponding to the gold secondary treatments (e.g. the half of a combined formula for an accompanying syndrome, or auxiliary herbs for a specific sign). A justified alternative path here means the treatment direction realized by the prescription differs from the gold standard but meets every condition of general principle (2).

- 5: same core path only, all conjunctive: (1) the sovereign-minister or principal herb groups cover the core treatment and all secondary elements; (2) the analysis explains the principal herbs one by one and corresponds fully to the prescription, with no key omission or function error; (3) the therapeutic focus is functionally equivalent to the gold principal formula and herbs for every gold secondary treatment are present (a different formula name, equivalent modifications, or herb substitutions do not affect this tier).

- 4: one of: (a) the focus functionally equivalent to the gold principal formula and realizing the core treatment (isolated weakly supported elements acceptable), the analysis consistent with the prescription, only one gold secondary herb unit missing; (b) a justified alternative treatment direction (general principle 2), the prescription fully realizing it and the analysis corresponding herb by herb, the discriminative-evidence field quoting the case text verbatim and the reason naming the core herb group.

- 3: one of: (a) equivalent to the gold principal formula but missing several secondary herb units, or only one of two coordinate cores realized, or the analysis leaving much unexplained or its role assignment slightly detached from the prescription; (b) a justified alternative path with flaws in realizing the treatment or partial detachment of the analysis; (c) an unjustified alternative path whose prescription still realizes the model's own treatment with the analysis corresponding.

- 2: one of: (a) an alternative path sustained by fabricated facts or by evading strong counter-evidence, or a prescription detached from its own treatment in several places; (b) regardless of the gold standard, the principal herb groups clearly deviate from the treatment focus, or the analysis is fragmented with several function errors.

- 1: the prescription essentially does not fit the treatment, or the analysis explains a different prescription (off-topic).

- 0: any of the three blocks missing, or the analysis completely detached from the prescription.

The output-format specification requires a single JSON object whose three keys are exactly the dimension names, each holding the situation (one of the three types, verbatim), the discriminative evidence (verbatim case quotations, or 'none'), the key counter-evidence ('none' if absent), an integer score of 0 to 5, and a one-sentence reason, under the same consistency constraints as the block-task variant.

## D. Human Evaluation Protocol

Each expert received two reading booklets and one scorecard. Booklet one (block-level content) presents, per case, the case text, the seven gold-standard blocks, and the seven blocks of each anonymized model; booklet two (logic consistency) presents the case text, the extractive reference reasoning chain, the relevant gold blocks, and the three dimension-specific block groups of each model. Model names, difficulty labels, and disease labels are hidden, and an automated leakage scan verifies their absence.

Five licensed TCM physicians scored independently in a single pass through an online questionnaire mirroring the scorecard. For each case and model, the scorecard collects ten integer ratings on an anchored 0-5 scale: seven content items judged against the gold-standard answer, then three logic-consistency items.

The ten items reuse definitions given earlier in this appendix: items 1-7 are the seven answer blocks as defined in the inference system prompt of Appendix A, and items 8-10 are the three logic-consistency dimensions as defined in the rubric of C.2. The anchors printed on the scorecard are as follows.

Content items (each block judged against the gold-standard answer): 5, core consistent with the gold standard and complete; 4, core consistent, only minor secondary omissions; 3, core consistent but with substantial omissions, or isolated errors; 2, core deviating or multiple errors; 1, essentially wrong or off-topic; 0, block not produced.

Logic-consistency items (consistency among the parts of the model's own answer, with reference to the reasoning logic of the gold standard): 5, fully coherent chain, consistent with the gold reasoning; 4, main line coherent, only isolated minor gaps; 3, main line holds but with clear gaps, or differs from the gold reasoning while internally consistent; 2, multiple breaks or clear flaws; 1, essentially contradictory or unrelated; 0, part missing, cannot be judged.

## E. Hyper-parameter Setup

### E.1 Candidate models

The five closed-source models were accessed through a single OpenAI-compatible client with identical messages and settings (Table E1) except the difference hyper-parameters derived from their official infrastructure: structured output enforced by strict JSON-Schema response formatting. Requests ran at 24 concurrent workers with 4 SDK-level and 3 application-level retries under exponential backoff, plus a guard that rejects and retries any response whose seven answer blocks are all empty.

Table E1. Closed-source candidate models and per-model settings.

| Model | Max new tokens |
|---|---|
| GPT-5.5 | 4,096 |
| Claude-Opus-4.8 | 4,096 |
| Gemini-3.1-Pro | 4,096 |
| Qwen3.7-Max | 4,592 |
| DeepSeek-V4-Pro | 4,096 |

For part of the benchmark, Qwen3.7-Max and DeepSeek-V4-Pro were accessed through their first-party APIs with equivalent settings: on DashScope, structured output requires the non-thinking mode; on the DeepSeek API, 4,096-token output cap applied, and the returned reasoning trace was taken as the Syndrome-Differentiation-and-Treatment Reasoning Path.

The two open-weight domain models were served locally with vLLM (Table E2): bfloat16 weights, GPU memory utilization 0.9, prefix caching, maximum context length 8,192 tokens, and schema-guided JSON decoding. Each model ran on one RTX 4090 48 GB GPU, with the whole dataset submitted as one continuous batch.

Table E2. Open-weight candidate models served locally with vLLM (one GPU per model).

| Model | Weights | Sampling | Output constraint | Max new tokens |
|---|---|---|---|---|
| ShizhenGPT-7B | 7B, bfloat16 | greedy (temperature 0, top-p 1.0) | guided JSON | 4,096 |
| HuatuoGPT-3-8B | 8B, bfloat16 | temperature 0.6, top-p 0.95, top-k 20 | guided JSON; thinking disabled | 4,096 |

## E.2 Judge models

Main judge (both tasks): Qwen3.7-Plus through the DashScope OpenAI-compatible API, temperature 0, JSON-object response format, thinking disabled, no output-length cap (a truncated response counts as an error and is retried). Judge calls ran at 8 concurrent workers with 3 application-level and 4 SDK-level retries.

Preliminary-experiment judges: the Qwen3.7-Plus side used the same settings as the main judge. The HuatuoGPT-3-32B side was served locally with vLLM at tensor parallel size 4 across four RTX 4090 48 GB GPUs (peer-to-peer transfers disabled for the consumer cards): thinking mode enabled, temperature 0.6, top-p 0.95, top-k 20, output caps of 8,192 tokens for the block task and 6,144 for the logic task (raised to 10,240 and 8,192 on retry, at most 2 retries), maximum context length 32,768 tokens, and a deterministic per-call seed derived by hashing the judge, model, case index, and attempt.

# F. API Cost and Inference Latency

## F.1 Inference cost

One full pass of the five closed-source models over the 2,000-case benchmark cost $353.32 at the account level, including retries and failed calls. Table F1 breaks the cost down by model from the measured mean token counts per case and the prices charged at run time (completion tokens include reasoning tokens); the breakdown totals $353.32. The two locally served models add no API cost.

Table F1. API inference cost of the closed-source models for one 2,000-case pass, computed from run-time prices (USD per million tokens) and measured mean tokens per case. Completion tokens include reasoning tokens.

| Model | Input price ($/M) | Output price ($/M) | Input tokens per case | Output tokens per case | Cost, 2,000 cases ($) |
|---|---|---|---|---|---|
| GPT-5.5 | 5 | 30 | 874 | 2,012 | 129.46 |
| Claude-Opus-4.8 | 5 | 25 | 1,477 | 2,539 | 141.72 |
| Gemini-3.1-Pro | 2 | 12 | 781 | 1,527 | 39.77 |
| Qwen3.7-Max | 1.25 | 3.75 | 595 | 4,481 | 35.09 |
| DeepSeek-V4-Pro | 0.7605 | 1.521 | 819 | 1,981 | 7.27 |

## F.2 Judging cost

One judge call scores all blocks or dimensions of one model answer, so a full pass over seven models is 14,000 calls per task. The main logic-chain pass is retrieval-heavy: 40.7k input tokens per call, 62% of them hitting the provider's prompt cache, and 225 output tokens; the measured six-model pass (11,982 calls) cost CNY 408 and the complete seven-model pass CNY 478. The block-level pass carries 52k input tokens per call and cost CNY 677. The retrieval-free preliminary prompts are an order of magnitude smaller (5.9k and 8.5k input tokens per call for the logic and block tasks); a full seven-model logic pass with the API judge cost CNY 125. The locally served 32B judge adds no API cost.

### F.3 Inference latency

API latency of the closed-source models was measured from the full-pass run log at 24 concurrent workers (Table F2): steady-state throughput ranges from 0.29 to 1.43 cases per second, and a serial pass over the five models takes 5.5 hours for the 2,000-case benchmark.

Table F2. Measured inference throughput and latency of the closed-source model APIs at 24 concurrent requests. Mean request latency is workers divided by throughput; wall time is for one 2,000-case pass.

| Model | Throughput (cases/s) | Mean request latency (s) | Wall time, 2,000 cases (min) |
|---|---|---|---|
| GPT-5.5 | 0.52 | 46 | 64 |
| Claude-Opus-4.8 | 0.43 | 56 | 78 |
| Gemini-3.1-Pro | 1.43 | 17 | 23 |
| Qwen3.7-Max | 0.29 | 83 | 115 |
| DeepSeek-V4-Pro | 0.64 | 38 | 52 |

The locally served candidates are far faster: with vLLM continuous batching on one RTX 4090, a 2,000-case pass for the 7B/8B models takes 20 to 40 minutes, one second of amortized latency per case. The API judge completes a full pass in hours: the six-model logic pass ran 11,982 calls in 3.2 hours at 8 concurrent workers (1.05 calls per second, 7.6 seconds per call), and the retrieval-free preliminary passes ran at a similar rate (13,811 calls in 3.4 hours). The local HuatuoGPT-3-32B judge is the slowest component: generating 2,000 thinking-mode tokens per call over 14,000 calls at tensor parallel size 4, a full pass takes 12 to 20 hours.

## G. Detailed Inter-Judge Agreement Results

Table G1 reports the full inter-judge agreement estimates underlying the analyses in judge-selection analyses in the main text.

Table G1. Case-level inter-judge agreement and systematic score offsets

| Assessment | Evaluated model | Huatuo judge − Qwen judge (percentage points) | Spearman ρ | CCC | Cases with absolute score difference ≤ 1 (%) |
|---|---|---|---|---|---|
| Block-level content | GPT-5.5 | +14.7 | 0.689 | 0.401 | 66.6% |
| | Qwen3.7-Max | +13.4 | 0.672 | 0.408 | 71.9% |
| | HuatuoGPT-3-8B | +20.4 | 0.668 | 0.303 | 47.6% |

| Assessment | Evaluated model | Huatuo judge − Qwen judge (percentage points) | Spearman ρ | CCC | Cases with absolute score difference ≤ 1 (%) |
|---|---|---|---|---|---|
| Cross-block logic consistency | GPT-5.5 | +8.2 | 0.465 | 0.359 | 91.5% |
| | Qwen3.7-Max | +11.2 | 0.360 | 0.278 | 78.5% |
| | HuatuoGPT-3-8B | +27.1 | 0.351 | 0.125 | 30.3% |

## H. Joint Metrics by Case Difficulty

Table H1 summarizes the cross-model mean scores and descriptive correlations within each difficulty stratum in the "Case-Level Association Between the Two Metrics" subsection.

Table H1. Cross-model mean content and consistency scores and descriptive mean correlations by difficulty

| Difficulty | Mean normalized content score | Mean normalized consistency score | Mean difference (percentage points) | Descriptive arithmetic mean of model-specific Pearson's r |
|---|---|---|---|---|
| Hard | 39.4% | 62.3% | 22.9 | 0.595 |
| Medium | 47.3% | 64.0% | 16.7 | 0.577 |
| Easy | 55.9% | 66.8% | 10.9 | 0.605 |

## I. Case Study

Aggregate scores establish relative model performance but do not show how individual verdicts are produced. We therefore examine one hard-tier case in detail, comparing the outputs of the seven evaluated models, the scores assigned by the three judge configurations, and the ratings of five physicians.

A hard follow-up case

This hard-tier case concerns a follow-up visit for irritable bowel syndrome. The patient reports one bowel movement daily, taking approximately five minutes, with scant, sticky stool and a persistent sense of incomplete evacuation. Additional symptoms include afternoon abdominal distension, cold hands and feet, postprandial hiccups, dry mouth with thirst, and a last menstrual period dated 19 March; appetite, sleep, and urination are normal.

Among the seven evaluation blocks, syndrome diagnosis, treatment-principles selection, and prescription generation are the decisive steps in clinical differentiation and the blocks for which the reference answer serves as the gold standard. For the remaining blocks, the reference represents one clinically admissible reasoning path among several. We therefore compare the three decisive blocks directly.

Table I1. Syndrome diagnoses, treatment-principles selections, and prescriptions generated by the seven evaluated models, with the reference answer shown for comparison. Prescriptions are reported as formula names and herb lists; preparation and administration instructions are omitted. All text was translated from the original Chinese.

| Model | Syndrome diagnosis | Treatment-principle selection | Prescription generation |
|---|---|---|---|
| Reference | Liver depression with spleen deficiency and damp-heat obstruction. | Soothe the liver and regulate the spleen, clear heat and transform damp, free the bowels and descend turbidity, with moistening of the intestines to ease defecation. | bupleurum 10 g, red peony 15 g, bitter orange 15 g, honey-fried licorice 5 g, bitter apricot kernel 10 g, hemp seed 15 g, magnolia bark 10 g, rhubarb 10 g, trichosanthes fruit 30 g, Chinese angelica 15 g, prepared pinellia 10 g, costus root 10 g, areca seed 10 g, radish seed 20 g, raw rehmannia 30 g, raw sanguisorba 15 g |
| GPT-5.5 | Spleen-yang insufficiency with damp obstructing the intestines; accompanying liver and stomach disharmony with stomach qi counterflow. | Warm and move spleen yang, strengthen the spleen and transform damp, regulate qi and move stagnation, harmonize the stomach and descend counterflow; moistening without cloying and freeing without harshness. | Xiangsha Liujunzi Tang plus Pingwei San, modified: codonopsis 10 g, stir-fried white atractylodes 12 g, poria 15 g, honey-fried licorice 6 g, tangerine peel 6 g, prepared pinellia 9 g, costus root 6 g, amomum fruit 5 g, black atractylodes 9 g, magnolia bark 9 g, agastache 9 g, eupatorium 9 g, bitter orange 9 g, chicken gizzard lining 10 g, dried ginger 3 g, fresh ginger 3 slices, jujube 3 pieces |
| Claude-Opus-4.8 | Spleen and stomach weakness with intermingled cold and heat and damp-induced qi stagnation; the cold extremities are ascribed to unroused spleen yang. | Open with acridity and descend with bitterness, harmonize cold and heat, strengthen the spleen and transform damp, regulate qi and descend counterflow. | Banxia Xiexin Tang, modified: prepared pinellia 12 g, coptis 5 g, scutellaria 9 g, dried ginger 9 g, codonopsis 15 g, honey-fried licorice 6 g, jujube 10 g, poria 20 g, stir-fried white atractylodes 15 g, tangerine peel 10 g, costus root 9 g, amomum fruit 6 g, raw coix seed 30 g, inula flower 10 g, hematite 15 g |
| Gemini-3.1-Pro | Spleen and stomach weakness with internal damp turbidity. | Strengthen the spleen and transform damp, regulate qi, harmonize the center and descend | Xiangsha Liujunzi Tang plus Pingwei San, modified: codonopsis 15 g, white atractylodes 15 g, poria 15 g, licorice 6 g, tangerine peel 10 g, pinellia 10 g, costus root |
| | | counterflow. | 6 g, amomum fruit 6 g, magnolia bark 10 g, black atractylodes 10 g, medicated leaven 15 g, ophiopogon 12 g |
| DeepSeek-V4-Pro | Spleen deficiency with damp stagnation and stomach qi counterflow. | Strengthen the spleen and dispel damp, regulate qi and relieve distension, harmonize the stomach and descend counterflow, with warming free flow and raising of the clear. | Xiangsha Liujunzi Tang plus Pingwei San and Juzhijiang Tang, modified: codonopsis 15 g, stir-fried white atractylodes 15 g, poria 15 g, honey-fried licorice 6 g, tangerine peel 10 g, ginger-processed pinellia 10 g, costus root 6 g, amomum fruit 6 g, magnolia bark 10 g, immature bitter orange 10 g, black atractylodes 10 g, cinnamon twig 10 g, pueraria 15 g, fresh ginger 3 slices |
| Qwen3.7-Max | Spleen deficiency with damp obstruction, qi stagnation and stomach counterflow; accompanying intermingled cold and heat and constrained yang qi. | Strengthen the spleen and transform damp, regulate qi and out-thrust yang, open with acridity and descend with bitterness, harmonize the stomach and descend counterflow. | Banxia Xiexin Tang plus Sini San, modified: prepared pinellia 10 g, scutellaria 10 g, coptis 3 g, dried ginger 6 g, codonopsis 15 g, bupleurum 10 g, white peony 15 g, immature bitter orange 10 g, magnolia bark 10 g, black atractylodes 10 g, inula flower 10 g, honey-fried licorice 6 g |
| ShizhenGPT-7B | Spleen-yang deficiency; accompanying qi stagnation with damp obstruction and stomach qi counterflow. | Warm yang and strengthen the spleen, move qi and transform damp, harmonize the stomach and descend counterflow. | Fuzi Lizhong Tang plus Pingwei San, modified: prepared aconite 6 g, dried ginger 9 g, codonopsis 15 g, stir-fried white atractylodes 12 g, poria 15 g, black atractylodes 9 g, magnolia bark 9 g, tangerine peel 9 g, bitter orange 12 g, inula flower 9 g, mume fruit 9 g, honey-fried licorice 6 g |
| HuatuoGPT-3-8B | Spleen and kidney yang deficiency with sinking central qi and stomach qi counterflow. | Warm and supplement spleen and kidney yang, raise central qi, descend counterflow and stop hiccups, dispel damp and stop diarrhea. | Fuzi Lizhong Tang plus Buzhong Yiqi Tang, modified: blast-fried aconite 10 g, dried ginger 10 g, codonopsis 15 g, white atractylodes 15 g, honey-fried licorice 10 g, astragalus 30 g, bupleurum 10 g, cimicifuga 10 g, tangerine peel 10 g, amomum fruit 6 g, hematite 30 g, clove 10 g, ophiopogon 15 g, schisandra 10 g |

No model identifies damp-heat. Six of the seven diagnose damp obstruction without heat, and four interpret the cold extremities as evidence of yang deficiency. Qwen3.7-Max alone attributes them to constrained yang qi, matching the reference interpretation. This diagnostic divergence propagates to treatment and prescription: none of the models includes rhubarb, hemp seed, or another bowel-freeing herb, whereas two select strongly warming formulas based on aconite and dried ginger. GPT-5.5 additionally excludes rhubarb and other purgatives because it infers possible pregnancy from the menstrual date, directly opposing the reference prescription. HuatuoGPT-3-8B departs furthest by adding kidney-yang deficiency and sinking central qi and receives the lowest physician scores for both block content and logic.

Table I2. Case-level scores. The main judge is Qwen3.7-Plus with retrieval grounding, the configuration used in the benchmark. Qwen internal and Huatuo internal denote Qwen3.7-Plus and HuatuoGPT-3-32B, respectively, scoring from internal knowledge without retrieval. Totals are reported after application of the counter-evidence cap, which reduced 10 of 14 Huatuo totals and none of the Qwen totals in this case. Expert scores are sums of the mean ratings assigned by five physicians across items. Block content is the sum of seven block scores (range, 0-35), and logic chain is the sum of three chain scores (range, 0-15).

| Model | Block content (0 to 35) | | | | Logic chain (0 to 15) | | | |
|---|---|---|---|---|---|---|---|---|
| | Main judge | Qwen internal | Huatuo internal | Experts | Main judge | Qwen internal | Huatuo internal | Experts |
| GPT-5.5 | 9 | 13 | 18 | 27.2 | 9 | 8 | 9 | 11.6 |
| Claude-Opus-4.8 | 6 | 22 | 21 | 21.8 | 9 | 12 | 13 | 10.6 |
| Gemini-3.1-Pro | 9 | 12 | 21 | 19.2 | 9 | 7 | 9 | 10.0 |
| DeepSeek-V4-Pro | 9 | 16 | 19 | 24.4 | 9 | 6 | 11 | 11.2 |
| Qwen3.7-Max | 13 | 23 | 19 | 22.8 | 9 | 12 | 12 | 11.2 |
| ShizhenGPT-7B | 7 | 12 | 19 | 19.8 | 9 | 8 | 11 | 10.0 |
| HuatuoGPT-3-8B | 7 | 11 | 12 | 15.4 | 8 | 7 | 7 | 7.6 |

The three judge configurations do not converge. For block content, the two internal configurations differ by as many as 9 of 35 points on the same answer and substantially alter the ordering: Gemini-3.1-Pro is joint second lowest under Qwen internal but joint highest under Huatuo internal. Their largest logic-chain difference is 5 of 15 points, for DeepSeek-V4-Pro. The main judge assigns 9 points to six of the seven models and 8 to HuatuoGPT-3-8B, providing almost no discrimination on logic. Its block-content scores range from 6 to 13, below the physician means of 15.4 to 27.2, consistent with the strictness offset observed in the human expert evaluation.

How the judge configurations reached their scores

To make these differences auditable, we trace the GPT-5.5 response along all three logic-chain dimensions. GPT-5.5 is the state-of-the-art model among those evaluated and the most representative subject: the physicians rate its answer highest among the seven models on both tasks of this case, at 27.2 of 35 for block content and 11.6 of 15 for the logic chain. Each table first shows the reference and model text of the blocks forming the link, then the record of each judge configuration.

Table I3. Scores assigned by the three judge configurations to the link from etiology and pathogenesis analysis to syndrome diagnosis in the GPT-5.5 response. The reference and model text of the two blocks forming the link are shown first. Verdict, differential evidence, key counter-evidence, and audit flags are recorded only by the internal-knowledge configurations. All text was translated from the original Chinese.

| | | | |
|---|---|---|---|
| Reference | Etiology and pathogenesis analysis: Spleen deficiency with qi stagnation is the root and internally accumulated damp-heat the manifestation. In the course of the illness, constrained liver qi invades the spleen crosswise, deranging the ascent and descent of the middle burner and congesting the qi dynamic, hence the distension and hiccups. Damp-heat pours down into the intestines and obstructs qi movement, hence the sticky stool, the incomplete evacuation, and the sagging sensation at the anus. The cold hands and feet are a sign of yang qi constrained and failing to reach the extremities. As yang qi wanes in the afternoon the stagnation worsens, so the distension is most evident then. The dry mouth and thirst arise from damp-heat blocking the distribution of fluids, with some insufficiency of the yin aspect.<br>Syndrome diagnosis: Liver depression with spleen deficiency and damp-heat obstruction. Main symptoms: sticky stool with incomplete evacuation, afternoon distension, cold hands and feet, hiccups after meals. Secondary symptoms: dry mouth with thirst, prolonged defecation, scant volume. Tongue and pulse: dark red tongue with yellow greasy coating, wiry slippery or deep rough pulse. | | |
| Model response | Etiology and pathogenesis analysis: The core pathogenesis lies in the middle burner. The spleen governs transportation and the stomach governs intake and descent; when spleen yang is insufficient, water and damp are not transported, turbid damp arises internally and pours into the intestines, hence the sticky, scant stool and the incomplete evacuation. Damp is sticky by nature, so the urge is present but the passage is unsatisfying. With spleen yang unaroused, clear yang fails to reach the extremities, hence the cold hands and feet. Damp and qi stagnation obstruct the middle burner, and as yang qi wanes in the afternoon the spleen works still more weakly, so the distension worsens then. After meals, if stomach qi fails to descend, liver qi invades crosswise, or the middle burner is blocked, qi rushes upward as hiccups. The dry mouth and thirst need not indicate excess heat; they can arise from damp obstructing the qi dynamic and the distribution of fluids, or from stomach qi failing to descend and carry fluids upward. Normal urination, appetite, and sleep suggest that the excess is mild and the deficiency primary. The overall pattern is root deficiency with branch excess: the root is insufficiency of spleen yang and spleen qi, the branch is turbid damp, qi stagnation, and stomach counterflow, located in the spleen, stomach, and intestines and involving the liver's free coursing.<br>Syndrome diagnosis: Main pattern: spleen-yang insufficiency with damp obstructing the intestines. Accompanying pattern: liver and stomach disharmony with stomach qi counterflow. Clinical basis: scant sticky stool with incomplete evacuation, afternoon distension, cold hands and feet, hiccups after meals, and dry mouth with thirst, with normal urination. | | |
| Field | Main judge | Qwen internal | Huatuo internal |
| Score (0 to 5) | 3 | 2 | 4 (capped to 3) |
| Verdict | Not recorded by this configuration | Alternative path, not admissible | Alternative path, admissible |
| Differential evidence | Not recorded by this configuration | None | Cold hands and feet and afternoon distension show spleen yang too weak to transport; dry mouth and thirst come from damp blocking fluid distribution rather than from heat. |
| Key counter-evidence | Not recorded by this configuration | The record holds no confirming yang-deficiency signs such as a pale tongue with white coating or a deep slow pulse, and sticky stool with incomplete evacuation usually points to damp-heat or qi stagnation rather than to spleen-yang deficiency alone. | A yellow greasy coating and a wiry slippery pulse are not reported, and spleen-yang deficiency has no typical support here such as aversion to cold or preference for warmth. |
| Audit flags | Not recorded by this configuration | None | Hard counter-contradiction; fabricated evidence |
| Rationale | Diverges from the reference core on an alternative path: the model gives spleen-yang deficiency with damp obstruction where the reference gives liver depression, spleen deficiency and damp-heat. The internal derivation is nonetheless self-consistent and matches the knowledge base. | The model forces the sticky stool onto spleen-yang deficiency with damp excess, passes over damp-heat and qi stagnation, and lacks the core tongue and pulse support for yang deficiency. The pathogenesis is supplied to close the argument. | Spleen-yang insufficiency with damp and qi stagnation explains the sticky stool, the distension and the cold extremities, and is distinguished from liver depression with damp-heat, which would show thirst with a yellow coating and loose stool with abdominal pain. |

The two internal configurations identify the same evidentiary gap but reach opposite verdicts. Both note that the case provides no tongue or pulse findings supporting yang deficiency. Qwen internal therefore classifies the alternative path as inadmissible and assigns 2 points; Huatuo internal classifies it as admissible and initially assigns 4. However, Huatuo internal simultaneously records a hard counter-contradiction and fabricated evidence, so the cap reduces the score to 3. Across the full case, Huatuo internal raises one or both flags in 35 of 49 audited cells, whereas Qwen internal raises none. The physicians nevertheless rate this GPT-5.5 dimension highest among the seven models, at 4.2 of 5.

Table I4. Scores assigned by the three judge configurations to the link from syndrome diagnosis to treatment-principle selection in the GPT-5.5 response. The reference and model text of the two blocks forming the link are shown first. Verdict, differential evidence, key counter-evidence, and audit flags are recorded only by the internal-knowledge configurations. All text was translated from the original Chinese.

<table>
<tr><td>Reference</td><td colspan="3">Syndrome diagnosis: Liver depression with spleen deficiency and damp-heat obstruction.<br>Main symptoms: sticky stool with incomplete evacuation, afternoon distension, cold hands and feet, hiccups after meals. Secondary symptoms: dry mouth with thirst, prolonged defecation, scant volume. Tongue and pulse: dark red tongue with yellow greasy coating, wiry slippery or deep rough pulse.<br>Treatment-principle selection: Soothe the liver and regulate the spleen, clear heat and transform damp, free the bowels and descend turbidity. Regulating the qi dynamic is the pivot, clearing damp-heat the assistant, with attention to moistening the intestines and easing defecation.</td></tr>
<tr><td>Model response</td><td colspan="3">Syndrome diagnosis: Main pattern: spleen-yang insufficiency with damp obstructing the intestines. Accompanying pattern: liver and stomach disharmony with stomach qi counterflow. Clinical basis: scant sticky stool with incomplete evacuation, afternoon distension, cold hands and feet, hiccups after meals, and dry mouth with thirst, with normal urination.<br>Treatment-principle selection: Warm and move spleen yang, strengthen the spleen and transform damp, regulate qi and move stagnation, harmonize the stomach and descend counterflow; moisten without cloying and free without harshness, so that intestinal qi flows without injury to the upright. If pregnancy is possible, avoid harsh purgatives and blood-breaking, fetus-disturbing agents and keep the treatment mild.</td></tr>
<tr><td>Field</td><td>Main judge</td><td>Qwen internal</td><td>Huatuo internal</td></tr>
<tr><td>Score (0 to 5)</td><td>3</td><td>3</td><td>4 (capped to 3)</td></tr>
<tr><td>Verdict</td><td>Not recorded by this configuration</td><td>Alternative path, not admissible</td><td>Alternative path, admissible</td></tr>
<tr><td>Differential evidence</td><td>Not recorded by this configuration</td><td>None</td><td>Postprandial hiccups and distension call for regulating qi and harmonizing the stomach; cold hands and feet call for warming yang rather than clearing heat.</td></tr>
<tr><td>Key counter-evidence</td><td>Not recorded by this configuration</td><td>The treatment is mainly warm and drying, which conflicts with the recorded dry mouth and thirst and with a possible damp-heat pathogenesis; no clearing of heat or generation of fluids is</td><td>No pregnancy caution is given although dried ginger and amomum are warm herbs.</td></tr>
</table>

| | | | |
|---|---|---|---|
| | | considered. | |
| Audit flags | Not recorded by this configuration | None | Hard counter-contradiction; fabricated evidence |
| Rationale | Diverges from the reference core treatment on an alternative path: the model advocates warming and moving spleen yang where the reference soothes the liver and clears heat. The model's treatment nonetheless corresponds fully to its own syndrome; the method follows the pattern. | The treatment follows the model's own diagnosis of spleen-yang deficiency with damp closely and is internally coherent, but because that diagnosis departs from key evidence in the case, such as the sticky stool and the dry mouth, the whole direction of treatment is off. | Warming and moving spleen yang with drying damp and moving qi covers spleen deficiency with damp and liver-stomach disharmony; it does not touch clearing heat or freeing the bowels, hence the difference from the reference. |

The pattern repeats on the link from syndrome diagnosis to treatment-principle selection. Qwen internal again rules the alternative path inadmissible, at 3 points, citing the conflict between the warm, drying treatment and the reported dry mouth and thirst. Huatuo internal again rules it admissible at 4 and is capped to 3. Its recorded counter-evidence states that the answer offers no pregnancy caution while using warm herbs; the treatment-principle selection block in fact states this caution explicitly, so the counter-evidence itself contradicts the answer it audits.

Table I5. Scores assigned by the three judge configurations to the link from treatment-principle selection through prescription generation to prescription analysis in the GPT-5.5 response. The reference and model text of the blocks forming the link are shown first; prescriptions are abbreviated, with the full herb lists given in the comparison table above. Verdict, differential evidence, key counter-evidence, and audit flags are recorded only by the internal-knowledge configurations. All text was translated from the original Chinese.

| | |
|---|---|
| Reference | Treatment-principle selection: Soothe the liver and regulate the spleen, clear heat and transform damp, free the bowels and descend turbidity. Regulating the qi dynamic is the pivot, clearing damp-heat the assistant, with attention to moistening the intestines and easing defecation.<br>Prescription generation: no named formula; 16 herbs built on rhubarb, hemp seed, trichosanthes fruit, and raw sanguisorba.<br>Prescription analysis: Soothing the liver and regulating qi: bupleurum relieves liver depression; bitter orange, costus root, and areca seed move qi and break stagnation, restoring the ascent and descent of liver and spleen. Clearing heat and transforming damp: trichosanthes clears damp-heat from lung and intestine; raw rehmannia and raw sanguisorba cool the blood, resolve toxin, and moisten dryness. Freeing the bowels and guiding out stagnation: rhubarb purges excess heat, hemp seed moistens the intestines, apricot kernel diffuses the lung and descends qi, magnolia bark and radish seed relieve distension and fullness. Harmonizing qi and blood: red peony quickens the blood and relaxes tension, Chinese angelica nourishes the |

| | | | |
|---|---|---|---|
| | blood and moistens the intestines, pinellia harmonizes the stomach and descends counterflow, and honey-fried licorice harmonizes the formula. | | |
| Model response | Treatment-principle selection: Warm and move spleen yang, strengthen the spleen and transform damp, regulate qi and move stagnation, harmonize the stomach and descend counterflow; moisten without cloying and free without harshness, so that intestinal qi flows without injury to the upright. If pregnancy is possible, avoid harsh purgatives and blood-breaking, fetus-disturbing agents and keep the treatment mild.<br>Prescription generation: Xiangsha Liujunzi Tang plus Pingwei San, modified; 17 herbs.<br>Prescription analysis: Codonopsis, stir-fried white atractylodes, poria, and honey-fried licorice carry the sense of Sijunzi Tang, strengthening spleen qi and supporting transportation as the root-treating group; once the spleen is strengthened, damp is moved and the sticky stool and incomplete evacuation should lessen. Tangerine peel and prepared pinellia dry damp, transform phlegm, regulate qi, and harmonize the stomach, restoring the ascent and descent of the middle burner for the distension and postprandial hiccups. Costus root and amomum move qi, arouse the spleen, and warm and harmonize the stomach, easing post-meal qi congestion and stomach counterflow; they are added late to preserve their aromatic nature. Black atractylodes and magnolia bark take the sense of Pingwei San, drying damp, moving the spleen, and relieving fullness, aimed at the sticky stool, distension, and stuffiness caused by damp in the middle burner. Agastache and eupatorium transform damp aromatically and arouse the spleen against turbidity, helping the dry, thirsty mouth without desire to drink and the failure of fluids to distribute. Bitter orange widens the middle and descends qi, guiding stagnation so that intestinal conduction runs freely and the scant, incomplete evacuation improves. Chicken gizzard lining disperses food and strengthens the stomach for the postprandial hiccups and weak transportation. Dried ginger and fresh ginger warm the middle, scatter cold, and harmonize the stomach with descent, with jujube harmonizing spleen and stomach, addressing the cold extremities and the insufficiency of spleen yang. The whole formula supplements while moving, warms while transforming, dries damp without harshness, and frees the bowels without attack, fitting spleen deficiency with damp, qi stagnation, and stomach counterflow. | | |
| Field | Main judge | Qwen internal | Huatuo internal |
| Score (0 to 5) | 3 | 3 | 4 (capped to 3) |
| Verdict | Not recorded by this configuration | Alternative path, not admissible | Alternative path, admissible |
| Differential evidence | Not recorded by this configuration | None | Xiangsha Liujun warms the spleen and boosts qi; black atractylodes and magnolia bark dry damp and relieve fullness; amomum and costus root move qi and harmonize |
| | | | the stomach. |
| Key counter-evidence | Not recorded by this configuration | The prescription consists entirely of warm, drying, spleen-strengthening, qi-regulating herbs, with no heat-clearing, damp-draining, or bowel-moistening herb, and cannot resolve the branch excess of sticky stool and incomplete evacuation. | There is no bowel-moistening herb, neither hemp seed nor rhubarb, so the incomplete evacuation is hard to resolve. |
| Audit flags | Not recorded by this configuration | None | Hard counter-contradiction; fabricated evidence |
| Rationale | The main line of the prescription diverges from the reference core on an alternative path, but the prescription realizes the model's own treatment principle, the prescription analysis matches the prescription, and the internal logic is self-consistent. | The prescription corresponds strictly to the model's treatment principle and the prescription analysis is clear, but it rests on a mistaken direction of differentiation and leaves the key symptoms of sticky stool and incomplete evacuation without treatment. | Sijunzi as sovereign with Pingwei San achieves warm transformation of damp stagnation and regulates qi to harmonize the stomach; the prescription analysis matches the treatment herb by herb but lacks intestine-moistening and bowel-freeing. |

On the link from treatment-principle selection through prescription generation to prescription analysis, the two internal configurations cite the same defect, the absence of rhubarb, hemp seed, or another bowel-freeing herb to resolve the incomplete evacuation, yet again reach opposite verdicts: 3 points against 4 capped to 3. After the caps, the Huatuo internal logic-chain total equals the main-judge total of 9 of 15; the agreement in totals conceals opposite reasoning.

## Discussion

The errors propagate coherently across the diagnostic and therapeutic blocks. Once damp-heat is omitted and the cold extremities are interpreted as yang deficiency, the treatment strategy shifts from clearing heat and freeing the bowels to warming yang and transforming damp, with corresponding changes in prescription selection. The disagreement between the internal judges is therefore not merely a difference in scoring severity: they identify similar evidential gaps but reach different conclusions about whether the alternative diagnostic pathway is admissible. Huatuo internal may accept a pathway while simultaneously flagging its supporting evidence as contradictory or fabricated. These findings demonstrate the value of retrieval-grounded scoring and explicit evidence traces for identifying inconsistencies among the assigned score, stated rationale, and audit record.